\documentclass{article}

\usepackage[letterpaper,top=2cm,bottom=2cm,left=2cm,right=2cm,marginparwidth=1.75cm]{geometry}
\usepackage[english]{babel}
\usepackage[autostyle]{csquotes}

\usepackage{amsmath,amssymb,amstext}
\usepackage{graphicx}
\usepackage{hyperref}
\hypersetup{
    unicode=false,          
    pdffitwindow=false,     
    pdfstartview={FitH},    
    pdfnewwindow=true,      
    colorlinks=true,        
    linkcolor=blue,         
    citecolor=green,        
    filecolor=magenta,      
    urlcolor=cyan           
}

\usepackage{braket}
\usepackage{amsthm}
\usepackage{bbold}
\usepackage{comment}

\title{Free Energy Projective Simulation (FEPS):  Active inference with interpretability}

\author{J. Pazem$^{1,*}$,  M. Krumm$^{1}$, A. Q. Vining$^{1,2}$, L. J. Fiderer$^{1}$, H. J. Briegel$^{1,3}$ \\
\textit{\small{$^{1}$ University of Innsbruck, Institute for Theoretical Physics, Technikerstraße 21a, A-6020 Innsbruck, Austria}} \\
\textit{\small{$^{2}$ Max Planck Institute of Animal Behavior, Department for the Ecology of Animal Societies,}} \\ \textit{\small{Bücklestraße 5a, 78467
 Konstanz, Germany}} \\
\textit{\small{$^{3}$ Department of Philosophy, University of Konstanz, Universitätsstraße 10, 78464 Konstanz, Germany}}\\
\small{$^{*}$ josephine.pazem@uibk.ac.at}}
\begin{document}

\maketitle
\begin{center}
    \begin{minipage}{0.9\textwidth}
{\small In the last decade, the free energy principle (FEP) and active inference (AIF) have achieved many successes connecting conceptual models of learning and cognition to mathematical models of perception and action. This effort is driven by a multidisciplinary interest in understanding aspects of self-organizing complex adaptive systems, including elements of agency. Various reinforcement learning (RL) models performing active inference have been proposed and trained on standard RL tasks using deep neural networks. Recent work has focused on improving such agents’ performance in complex environments by incorporating the latest machine learning techniques. In this paper, we take an alternative approach. Within the constraints imposed by the FEP and AIF, we attempt to model agents in an interpretable way without deep neural networks by introducing Free Energy Projective Simulation (FEPS). Using internal rewards only, FEPS agents build a representation of their partially observable environments with which they interact. Following AIF, the policy to achieve a given task is derived from this world model by minimizing the expected free energy. Leveraging the interpretability of the model, techniques are introduced to deal with long-term goals and reduce prediction errors caused by erroneous hidden state estimation. We test the FEPS model on two RL environments inspired from behavioral biology: a timed response task and a navigation task in a partially observable grid. Our results show that FEPS agents fully resolve the ambiguity of both environments by appropriately contextualizing their observations based on prediction accuracy only. In addition, they infer optimal policies flexibly for any target observation in the environment.}
    \end{minipage}
\end{center}
\vspace{1cm}

\section{Introduction}

A key challenge in both cognitive science and artificial intelligence is understanding cognitive processes in biological systems and applying this knowledge to the development of artificial agents. In this work, we develop an interpretable artificial agent which integrates key aspects of both reinforcement learning (RL) \cite{SuttonBarto1998} and active inference \cite{Kirchhoff2018, Linson_Clark_Friston2018, Pezzulo2015_homeostasis, Raja2021, Mazzaglia2022}. In RL, an external reward signal is typically used to guide an agent’s behavior while in active inference no such reward signal exists. Instead, agents follow an intrinsic motivation which is rooted in the free energy principle (FEP) \cite{Tschantz2020,Friston_AIF_and_learning2016}. 

Central to the FEP is the idea that adaptive systems can be modeled as performing an approximate form of Bayesian inference. The outcome of this process minimizes a quantity, called variational free energy (VFE), which lends its name to the FEP. The FEP encompasses various paradigms that align with the goal of Bayesian inference, such as predictive processing \cite{Bubic2010} and the Bayesian brain hypothesis \cite{Vilares2011}.

One way to implement the FEP is through active inference which involves a planning method for action selection based on an internal model of the environment, referred to as the world model, along with a preference distribution, which encodes the agent’s desired states (e.g., maintaining a certain body temperature). According to the FEP, it is—in principle—always possible to explain the observed behavior of living systems through active inference \cite{Friston_2010_unified_brain}. The Free Energy Principle has been applied to a wide range of domains, including neuroscience \cite{Friston_2010_unified_brain, Friston_Predictive_coding_FEP_2009, Pezzulo_Neural_representation_in_AIF_2024}, psychology \cite{Cullen2018,McGovern2022_FEP_Anxiety}, behavior biology \cite{Ramstead_Schrodinger_question_2018, Heins_collective_motion_2024, Pezzulo2015_homeostasis}, and machine learning \cite{Mazzaglia2021, Fountas2020, Nguyen2024_robustAIF, Tschantz_RLthroughAIF_2020, DeTinguy2024}.



Active inference has recently gained traction in theoretical discussions \cite{Raja2021, Clark2013, Clark2017,Kagan_AIFexp2022, Smith2021} leading to several successful algorithmic implementations \cite{Mazzaglia2022, Nguyen2024_robustAIF, Mazzaglia2021,Cullen2018,DeTinguy2024, Tschantz2020, Lanillos2021,Fountas2020,Piriyakulkij2023}. Existing implementations are based on methods developed in the context of model-based reinforcement learning and rely mostly on neural networks to represent the world model \cite{Cullen2018,Tschantz2020,Mazzaglia2021, Mazzaglia2022, DeTinguy2024, Kawahara2022, Tinker2024, Lanillos2021}. As an alternative to neural networks, this work proposes implementing active inference using Projective Simulation (PS) \cite{Briegel2012}.

In PS, memory is organized as a directed graph, which—unlike neural networks—is designed to be interpretable such that individual vertices carry semantic information \cite{Ried_Eva2023, Flamini2023}. The agent’s cognitive processes are modelled as (1) a random walk through the graph (deliberation), and (2) the updating of transition probabilities along the graph’s edges (learning). The primary motivation for using PS over neural networks is its inherent interpretability.

While so far PS has mostly been employed in the RL framework, this work extends PS to the active inference framework by proposing and testing an agent model called Free Energy Projective Simulation (FEPS). FEPS is an active inference model which uses PS for both the world model and the action policy, combining existing and novel methods for training these components. Key features of FEPS include an internal reward signal derived from the prediction accuracy, and the use of two distinct preference distributions—one for learning the world model and another for achieving a specific goal. For the latter purpose, we propose new heuristics to estimate the long-term value of belief states in sight of the goal.

In the following sections, we first introduce active inference (Section \ref{sec: FEP}) and PS (Section \ref{sec: PS}), before presenting the architecture (Section \ref{sec: FEPS architecture}) and algorithmic features (Section \ref{sec: FEPS algo features}) of the FEPS model. We then test FEPS through numerical simulations in biologically inspired environments, focusing on timed responses and navigation (Section \ref{sec: numerical results}). Finally, we conclude with a discussion of future research directions (Section \ref{sec: Discussion}).

\section{Active Inference} \label{sec: FEP}

The following presentation of active inference is divided into two parts. First, we outline how the world model is constructed and updated in response to sensations (perceptual inference). Next, we explain how planning and action selection are modeled (active inference).

During perceptual inference, the agent develops the world model \cite{Ha2018, Mendonca2021}, sometimes referred to as a generative model, highlighting its ability to generate simulated percepts.  The world model \cite{Friston_2010_unified_brain, Mazzaglia2022},  is described as a partially observable Markov decision process (POMDP) ($\mathcal{B}, \mathcal{S}, \mathcal{A}, T, L, R$), with belief states $\mathcal{B}$, observations $\mathcal{S}$, and actions $\mathcal{A}$. The \textit{transition function} $T$ assigns probabilities to transitions between belief states given some action, while $L$, sometimes referred to as the \textit{emission function}, defines the probability that some belief state $b\in\mathcal{B}$ emits a sensory signal $s\in\mathcal{S}$. Note that belief states are observable to the agent and that they are generally different from the actual, unobservable, states of the environment.

In what follows, the time-indexed random variables $B_t$, $S_t$, and $A_t$ represent beliefs, observations, and actions at a given time $t$, with their possible values denoted by the corresponding lower-case letters.
 The transition function $T$ between times $t-1$ and $t$ corresponds to the conditional distribution $p(B_{t}|B_{t-1}, A_{t-1})$, while the emission function $L$ at time $t$ is the likelihood $p(S_t|B_t)$. Finally, rewards $R$ are used to learn the transition and emission functions. In this work, rewards are internal (generated by the agent itself) \cite{Oudeyer2007} but in general they could also be external (given by the environment). 
It is further assumed that action $A_t$ is selected based on the current belief state $B_t$, as represented by the conditional distribution $\pi(A_t|B_t)$, which defines the policy. Combining these elements, the joint distribution of the world model up to time $t$ can be decomposed as follows:

\begin{equation} \label{eq: world model}
p( B_{0:t},A_{0:t-1}, S_{0:t}) = p(B_0, S_0) \prod_{t'= 1} ^{t}  \pi(A_{t'-1}|B_{t'-1})\, p(S_{t'}|B_{t'}) \,p(B_{t'}|B_{t'-1}, A_{t'-1}),
\end{equation}
where the index notation $B_{0:t}=(B_0,B_1,\dotsc, B_t)$ is used for sequences of random variables, and $p(B_0, S_0)$ denotes the initial distribution over beliefs and observations. Equation \eqref{eq: world model} defines the relations between random variables: while the distribution over observation and an action is specified by the current belief state, both the previous belief state and the previous action determine the current belief state.

The transition function $p(B_t|B_{t-1},A_{t-1})$ plays a crucial role in active inference because it has to be learned by the agent. Following the idea of active inference, learning involves updating $p(B_t|B_{t-1},A_{t-1})$ through variational inference, an approximate version of Bayesian inference. Variational inference simplifies Bayesian inference by restricting the set of possible posterior transition functions to a family $\{Q_\phi(B_t|B_{t-1},A_{t-1})\}_\phi$, parametrized by some variable $\phi$, thus reducing computational complexity. When a new observation $s_t^\text{env}$ is received from the environment, the approximate posterior is obtained as the solution to an optimization problem which minimizes the VFE, here conditioned on the specific values $b_{t-1}$ and $a_{t-1}$ for the random variables $B_{t-1}$ and $A_{t-1}$ in the previous step:
\begin{equation} \label{eq: VFE}
\mathcal{F} = \ \mathbb{D}_{\mathrm{KL}} \left[q_\phi(B_t | b_{t-1}, a_{t-1}) || p(B_t | b_{t-1}, a_{t-1}) \right]
+  \mathbb{E}_{b_t \sim q_\phi(B_t | b_{t-1}, a_{t-1})} \left[-\log(p(s_t^\text{env} | b_t)\right], 
\end{equation}
where $\mathbb{E}_{b_t \sim q_\phi(B_t | b_{t-1}, a_{t-1})}[f(b_t)] = \sum_{b_t} q_\phi(b_t | b_{t-1}, a_{t-1})\, f(b_t)$ represents the expectation value over belief states distributed according to the posterior distribution of some function $f$ of $b_t$. Minimizing the VFE balances two effects: while the first term penalizes drastic changes in the distribution, the second promotes accuracy in the model to predict observations coming from the environment. The Kullback-Leibler distance $\mathbb{D}_{\mathrm{KL}}(\bullet)$ represents the dissimilarity between the posterior and prior distributions for the transition function, $q_\phi(B_t | b_{t-1}, a_{t-1})$ and $p(B_t | b_{t-1}, a_{t-1})$ respectively. The second term is the surprise of the current observation. In order to calculate this surprise, an expectation value is calculated from the posterior distribution $q_\phi(B_t | b_{t-1}, a_{t-1})$ over belief states. Realizing that the first term is either larger or equal to zero, the VFE is an upper bound on the surprise raised by the current input from the environment. By updating its model and minimizing the variational free energy, the agent therefore minimizes its surprise. Typically, the prior is replaced by the posterior after collecting a number of observations and implementing the corresponding updates on the posterior.

So far, the model acquisition has been described in terms of perceptual inference: the system has no control over its actions yet, and constructs the internal model based solely on its observations. The distribution over actions, or policy $\pi(A_t|B_t)$, can take any form and is not yet optimized over.

A FEPS agent performs active inference \cite{Kirchhoff2018} by exploiting the world model that has been learned through perceptual inference in order to plan sequences of future actions. Using the current world model, the agent estimates the future free energy for each of its actions based on the current transition function and implements the one action that minimizes it. This estimate of the free energy for future states does not have a unique form \cite{Millidge2020} and different formulations can lead to different trade-offs between exploration and exploitation. Here, we use the most common one, denoted in the literature as the \textit{Expected Free Energy} (EFE):
\begin{align}
\mathcal{G}_{b_{t-1}}[a]
&= \mathbb{E}_ {b_t,s_t \sim p(B_t , S_t| b_{t-1}, a)} [\log  p(b_t |b_{t-1}, a)
 - \log \text{pref}(s_t, b_t | b_{t-1},a)], \\
&= - \mathbf{H}[B_t|b_{t-1},a] + \mathbb{E}_{s_t,b_t \sim p(S_t,B_t|b_{t-1}, a)}[\mathcal{S}^\text{pref}(s_t, b_t|b_{t-1}, a)] \label{eq: EFE}
\end{align}
where $p(B_t , S_t| b_{t-1}, a)$ refers to the world model over a single time step, and the surprise of getting outcomes $s_t$ and $b_t$ is denoted $\mathcal{S}^\text{pref}(s_t,b_t|b_{t-1}, a) = -\log \text{pref}(s_t,b_t|b_{t-1}, a)$. $\mathbf{H}(Y|x) = -\sum_y p(Y=y|X=x) \log p(Y=y|X=x)$ stands for the conditional entropy of the random variable $Y$ conditioned on a specific value $X=x$. It corresponds to the expected surprise over belief states and is therefore always positive. $\text{pref}(S_t, B_t | b_{t-1},a)$ is a distribution over belief states and observations.

The EFE relies on two measures to determine actions: how much uncertainty is expected, and how useful is the action in order to fulfill a goal. In an active inference setting, an action's utility refers to its expected capacity to meet some preferences. The first term in Equation \eqref{eq: EFE} is the negative entropy over belief states, related to the expected information gain about the transition function: the larger the entropy, the larger the gain. Maximizing this entropy minimizes the expected free energy and leads to explorative behaviors \cite{Millidge2020}.  Finite entropy can have two causes: (1) the transition to the next state is still uncertain in the world model, or (2) the transition to the next state in the environment is stochastic. In the second case, if entropy were used alone, agents could fall in the so-called curiosity trap \cite{Pathak2017} by always choosing actions that lead to random outcomes in the environment and inherently result in high entropy values. The second term reflects the utility gained from taking the action under consideration. When minimized, the transitions are expected to match the preferences to a greater extent. Therefore, minimizing the EFE maximizes the utility an agent expects from its action. For this purpose, a new biased distribution, the preference distribution $\text{pref}(S_t, B_t | b_{t-1},a)$, encodes the desirability of some states and observations. The larger the preference for a state, the larger the probability in the preference distribution, and the smaller the associated surprise. This second term encourages exploitation of the world model in order to fulfill the preferences. 
In a biological agent, these preferences could be of genetic origin (a preference for homeostatic states for example), socially learned (a preference for some type of songs for birds of certain areas, that would be different for the same species in a different place), acquired or externally given. As a result, they can encode a set of states and observations that are favorable to the survival of the agent. As a modeling choice, preferences cover both belief states and observations, and are conditioned on the previous state and action.

For the purpose of learning with an artificial agent, the preferences can naturally encode a task to fulfill as a preferred target state or observation. Choosing the action that minimizes the EFE, the agent selects the transition in its world model that will bring it as close as possible to its preferred states according to its world model. Furthermore, by using the full world model over long periods of time, say from 0 to $t$, as in Equation \eqref{eq: world model}, the long-term EFE can be calculated to project $t$ steps ahead in the world model and plan for sequences of $t$ actions. Finding the optimal sequence of actions can rely on tree search with pruning \cite{Da_Costa_MC_2020}, or calculating a long-term, discounted, expected free energy quantity \cite{Paul2024, Friston_sophisticated_2021, Kawahara2022} for example.

To summarize, according to the free energy principle, an agent that adapts to its surroundings can be modeled as learning a world model of its environment. In this process, the variational free energy is minimized, reflecting the reduction in the surprise the agent experiences about new sensory events. Active inference prescribes a method to plan actions in order to reduce this surprise. In other words, actions are chosen to consolidate the world model. An action is chosen if it minimizes the expected free energy, or equivalently, if it is expected to lead to transitions associated with high uncertainty and high utility. As noticed already in \cite{Millidge2020}, this can be counter-intuitive at first glance, since minimizing the VFE and the EFE respectively minimizes and maximizes uncertainty. This paradox can be resolved by realizing that in active inference, actions target areas in the world model whose outcomes are least predictable in order learn about them and to receive less surprising outcomes in the future, thereby minimizing the VFE in the long run.

\section{Projective Simulation} \label{sec: PS}

Projective Simulation (PS) \cite{Briegel2012} is a model for embodied reinforcement learning and agency inspired from physics that performs associative learning on a memory encoded as a graph. It is composed of a network of vertices, denoted clips, with a defined structure that gives each clip a semantic meaning, that can be assigned from the start or acquired progressively through past experiences. For example, a clip may represent a sensory state the agent's sensors are capable of receiving, or it can inherit supplementary semantics from past experiences and reinforcements. Directed edges between clips are weighted and can be modified dynamically to learn and adapt to the environment. In particular, PS allows the simulation of percept-action loops to represent previous interactions with the environment and the associated decision process. The resulting graph is the Episodic and Compositional Memory (ECM). When a clip is visited, either because it is currently experienced, or because it is used for deliberation, it is excited. After a first clip was excited, deliberation takes place as a traceable random walk in the ECM originating from this initial clip. It ends when a decoupling criterion is met, which leads to an action on the environment.

For the purpose of solving different environments while retaining interpretability, the ECM can adopt different structures. In order to imitate a percept-action loop, the ECMs are often structured as bipartite graphs, where a first layer contains percepts and the second is composed of actions \cite{Melnikov2014, Mautner2015}. In this case, the trained ECM is directly relatable to a policy. For more complex environments, or in order to extract abstract concepts from the percepts, an intermediate layer can be added between the percept and action layers \cite{Eva2023, Melnikov2017}. The ECM can either contain a fixed number of clips, or it can dynamically add and erase some of them when needed \cite{Melnikov2017}. To consider composite percepts and actions, the ECM graph can be replaced by a hypergraph, where each hyperedge connects a set of clips to another set \cite{Lemaitre2024}.

Each directed edge is equipped with at least one attribute to track learning and allow the agent to react adaptively to the environment. $h$-values increase as the edges are rewarded. They encode the strength of associations between clips that are useful to fulfill some task and record the learning of the agent. The probability of a transition associated with an edge with $h$-value $h_{ij}$ connecting two clips,  $c_i \rightarrow c_j$, is inferred from the $h$-values:
\begin{equation} 
p(c_j|c_i) = \frac{h_{ij}}{\sum_{k} h_{ik}}.
\end{equation}
Alternatively, a softmax function can also be implemented to enhance the differences between probabilities, especially in large ECMs. Upon receiving a new percept, the agent deliberates by taking a random walk through the ECM, using the probabilities defined from the $h$-values to sample a new edge.

As the agent modelled with PS interacts with its environment, it receives rewards that are distributed over the different edges, which changes the corresponding $h$-values. Specifically, the $h$-value of the edge $c_i \rightarrow c_j$ is updated as follows:
\begin{equation}
\label{eq: PS update}
h^{t+1}_{ij} = h^{t}_{ij} - \gamma(h^{t}_{ij} - h^{0}_{ij}) + R 
\end{equation}
where $\gamma$ is the forgetting parameter, $h_{ij}^0$ is the initial $h$-value for  $c_i \rightarrow c_j$ and $R$ is the reward. When the reward is positive, the $h$-value of the corresponding edge is increased accordingly. If an edge is not visited or does not receive a positive reward, the corresponding $h$-value decreases back to their initial value thanks to the forgetting mechanism in the second term.  In order to accept continuous percepts and to unlock generalization on some task, neural networks have been used to update the $h$-values in some cases. Training was then implemented by minimizing a loss function \cite{Jerbi2021}.

Projective simulation has been tested in multiple tasks, ranging from the standard RL toy environments \cite{Melnikov2014,Mautner2015}, robotics \cite{Hangl2020}, to animal behavior simulations \cite{LopezIncera2020, LopezIncera2021} and quantum computations \cite{Tiersch2015,Trenkwalder2022}. Extension of the model include modeling the ECM with a quantum photonic circuit \cite{Flamini2023} and considering composite concepts in the form of multiple joint excited clips using hypergraphs \cite{Lemaitre2024}.

\section{The Free Energy Projective Simulation agent} \label{sec: FEPS architecture}

We combine Projective Simulation, with Active Inference, following the free energy principle's framework. A FEPS agent is a model-based Projective Simulation agent, where the world model is an ECM with a clone-structured architecture \cite{Dedieu2019, Guntupalli2023}. Consequently, clone clips inherit the semantics of the unique sensory state they relate to, and context creates distinctions between hidden states that emit the same observations. As in the FEP, the agent does not need external rewards. Instead, prediction accuracy, weighted with confidence, is used as a reinforcement signal. The world model is directly exploited by the agent to set the edges' $h$-values in the policy with active inference.

\subsection{Architecture of the agent} \label{subsec: FEPS architecture}

\begin{figure*}[t!]
    \centering
    \includegraphics[width=1\linewidth]{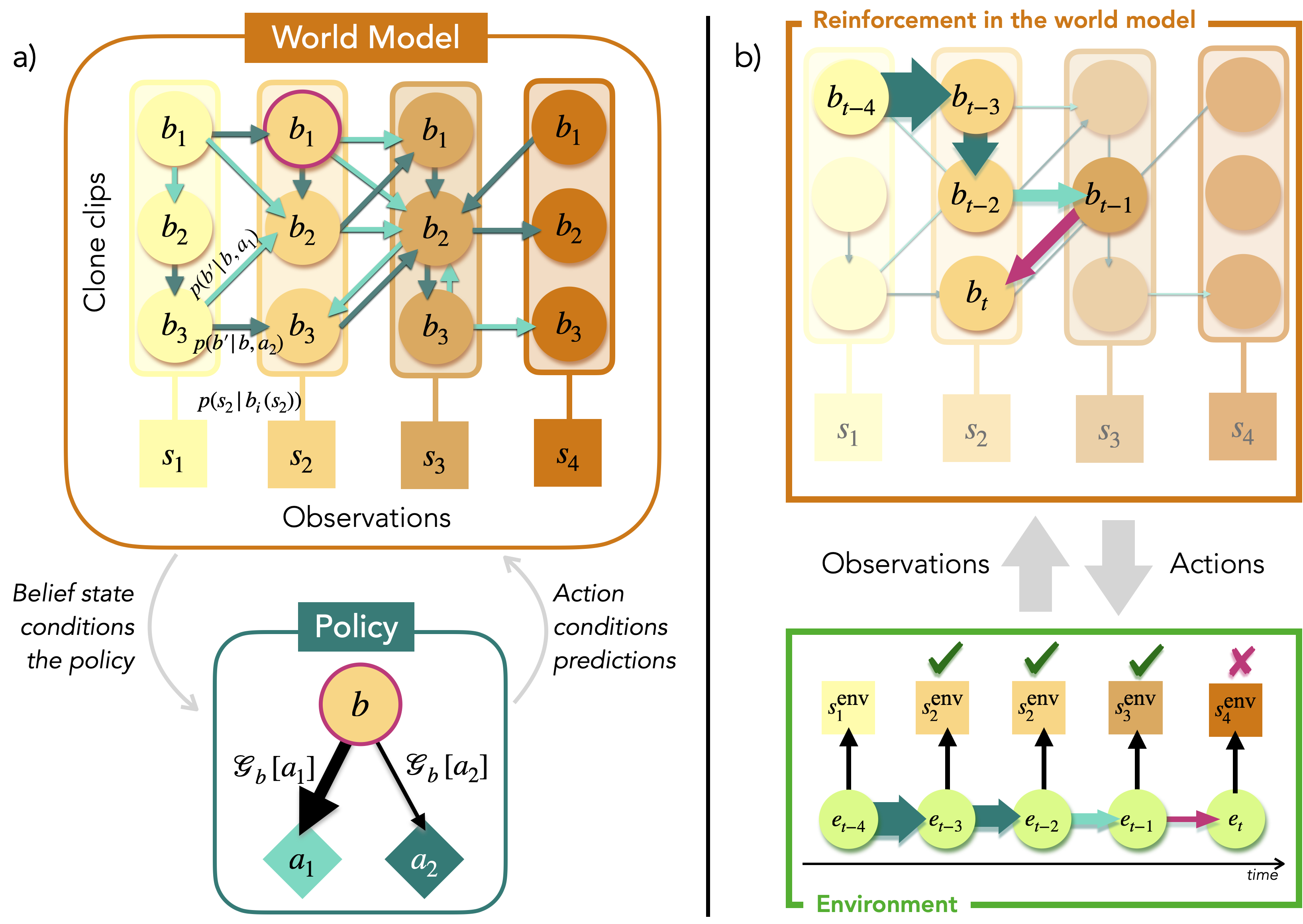}
    \caption{\textbf{Architecture and training of an FEPS agent} a) Architecture of a FEPS agent, with four sensory states (squares) and two possible actions (diamonds). The agent has two main components: the world model and the policy. The world model is composed of vertices representing observations (squares) while clone clips represent all values a belief state can take (circles). As in a clone-structured graph, each clone clip $b$ relates to exactly one observation $s$ and the emission function $p(s|b)$ is deterministic. The clone clips, together with the set of edges between them, form an ECM. A belief state, circled in purple, is designated by an excited clone clip. The weighted edges in the ECM encode the transition function and are trainable with reinforcement: there is one set of edges per action (light and dark turquoise arrows). The belief state in the ECM is an input to the policy, where the probability of sampling an action is a function of the EFE. In turn, the action that was selected determines the edge set to sample from in the world model in order to make a prediction for the next belief state and observation. 
    b)  Training of the world model of a FEPS agent. The agent interacts with the environment by receiving observations and implementing actions. When an action $a_{t}$ is chosen, a corresponding edge $b_{t} \xrightarrow{a_{t}} b_{t+1}$ is sampled in the world model, from the current to the next belief state, conditioned on the action. The observation $s_{t+1}$ associated with the next belief state is the prediction for the next sensory state. Simultaneously, the action is applied to the environment and creates a transition in the hidden states of the environment, $e_{t} \xrightarrow{a_{t}} e_{t+1}$ (bottom, green rectangle). This transition is perceived by the agent through the observation $s^\mathrm{env}_{t+1}$. Finally, the weights of the edges are updated. The reinforcement of an edge is proportional to the number of correct predictions it enabled in a row, as depicted with the thickness of the arrows in the world model. When the agent makes an incorrect prediction (the purple arrow),  the reinforcements are applied to the edges that contributed to the trajectory. The last, incorrect, edge is not reinforced.}
    \label{fig:FEPS model}
\end{figure*}

To mimic a system described by the FEP, the agent is composed of two structures: the world model and the policy, each represented by separate graphs with vertices corresponding to random variables, and edges that can be sampled to perform a random walk, as in figure \ref{fig:FEPS model}. Consider the agent can perceive $N_S$ sensory states, has $N_B$ possible belief states and a repertoire of $N_A$ actions. Each state is supported by one vertex in a graph. The world model's vertices can either support belief states or sensory states. The world model is the representation of the environment (see Eq. \ref{eq: world model}) required by active inference. For reasons that will become clear shortly, the ECM of a FEPS agent is made of all vertices, that we call clips, that support belief states, and edges that represent transitions between such clips. A belief state is then formally defined by the excitation configuration of clips at one step of the deliberation. We limit the number of excitations in the ECM at any given time to one. Such a vertex deserves the name "clip" because it receives an excitation when the corresponding representation of an event is revisited in order to make a decision. The policy covers all possible conditioned responses coming from any belief state, given the repertoire of actions of the agent.

In the world model, two sets of edges can be traversed at different times during the deliberation: we denote them emission and transition edges respectively. They aim at predicting and explaining sensory signals received from the environment.

In the first set, emission edges $b \rightarrow s$ relate belief states and sensory states, modeling the latter as parts of larger, possibly contextualized, hidden states by using clone-structured Hidden Markov Models (HMM)  \cite{Dedieu2019, George2021, Guntupalli2023}. Each clip is bound to a single observation and is denoted a \textit{clone clip}. A single edge in the emission set carries a non-zero probability for each clip, as shown in figure \ref{fig:FEPS model}a. Consequently, this set of edges defines a deterministic likelihood $p(s_t | b_t) = \delta_{s_t, s(b_t)}$ in the world model in Eq. \eqref{eq: world model} and the agent remains initially agnostic of the dynamical structure of the hidden states $\{e_{\tau}\}_{\tau=0}^t$ in the environment. Meanwhile, a clone clip can readily be interpreted as an augmented description of a specific observation. In particular, the additional information can relate to the cause or the context of the observation, such as the previous belief state and previous action, for example. 
Sampling some sensory state $\hat{s}_{t+1}$ amounts to predicting the next observation: if it turns out to coincide with the actual observation perceived from the environment, a reward will be distributed to the edges contributing to the random walk that led to this prediction.  For clarity, we denote predicted states with hatted low case letters. We choose to associate each observation to a fixed number $N_\text{clones}$ of clone clips, such that  $N_B = N_S \times N_\text{clones}$. This approach works remarkably well for navigation problems \cite{George2021, Guntupalli2023}. Transferring the dynamics learned on some set of sensory states to another set is also possible by keeping the transition function unchanged, but redistributing clone clips to the sensory states in the new set \cite{Guntupalli2023}.

The purpose of the set of transition edges $b_t \xrightarrow{a_t} b_{t+1}$ in the world model is to encode the presumed dynamics in the environment as transitions between belief states, conditioned on actions. They are represented as edges between clone clips in figure \ref{fig:FEPS model}a. In contrast to other sets of edges, transition edges are endorsed with attributes such as $h$-values that enable learning with reinforcements (as in Eq. \ref{eq: PS update} for example). Therefore, clone clips together with transition edges constitute an ECM for the FEPS agent. From a given clone clip, for each action, a set of edges points to the next possible estimated belief states. The $h$-values of those edges indicate how certain the agent is that taking a particular action from the current belief state will lead to any of the clone clips in the future. There can be at most $N_B$ edges in each such group of transition edges. The full set of transition edges defines the transition function $p(B_{t+1}|B_t, A_t)$ in the world model in Eq. \eqref{eq: world model} and corresponds to the trained part of the model. Before reinforcement, this distribution is referred to as the prior. The posterior is the updated version of the transition function, after the agent distributed rewards to the relevant transition edges. The posterior is labeled $q(B_{t+1}|B_t, A_t)$. 

The final component of the agent guides its behavior. The policy $\pi(A_t|B_t)$ is modeled as a separate graph with two layers of $N_B$ and $N_A$ clips respectively. Given the clone clip corresponding to belief state $b_t$ is excited in the world model, an action $a_t$ is sampled to be applied to the environment, based on how much surprise it expects from this decision. Each edge $b_t \rightarrow a_t$ is weighted with the expected free energy $\mathcal{G}_{b_t}[a_t]$, that defines the policy. 

\subsection{Reinforcement with prediction accuracy}

Each interaction step with the environment involves a deliberation over three states: (1) the next belief state is be proposed, (2) the next sensory state is predicted and (3) an action is chosen. The agent excites a belief state $b_{t+1}$ it believes it will transition to, given its current action $a_t$, by sampling a transition edge $b_t \xrightarrow{a_t} b_{t+1}$. From there, the agent makes a prediction $\hat{s}_{t+1}$ about the next sensory state. Meanwhile, an action $a_{t+1}$ is selected in the policy and it is applied to the hidden state in the environment, that emits a new observation. The interaction step ends by comparing the predicted and perceived sensory states, $\hat{s}_{t+1}$ and $s_{t+1}^\text{env}$. 

The world model is trained without external rewards, and reinforcement is instead based on matching predictions and observations. We call \textit{trajectory} a sequence of transitions that led to correct predictions about sensory states. To record the trajectory, transition edges are equipped with a new attribute: the \textit{confidence}, $f$. Initialized at zero, it increases for all transitions in the trajectory every time the prediction $\hat{s}_{t+1}$ and the actual sensory state $s^\text{env}_{t+1}$ coincide. The more subsequent predictions an edge enabled, the higher the confidence for that edge: it reflects the number of correct predictions the edge enabled until the end of the trajectory. Formally, a trajectory $\tau$ is a sequence of transitions whose predictions were confirmed by the observations from the environment. If at step $n$ in the $t$-th trajectory $\tau_t$ the sensory prediction was accurate, confidence is enhanced for all edges $i\rightarrow j$ in $\tau_t$:

\begin{equation}
f_{ij}^{(\tau_t), n} = \begin{cases}   
f_{ij}^{(\tau_t), n-1} + 1    & \text{if}\ b_i \rightarrow  b_j \in \tau_t \\
0 & \text{otherwise.}
\end{cases}\end{equation}

When the prediction and observation do not match, the trajectory $\tau_t$ is interrupted, and the rewards are distributed to the transition edges' $h$-values proportionally to the corresponding confidence:
\begin{equation}
\label{eq: FEPS update}
h^{t}_{ij} = h^{t-1}_{ij} - \gamma(h^{t-1}_{ij} - h^{0}_{ij}) + f_{ij}^{(\tau_{t})} R 
\end{equation}
where $h_{ij}^{t-1}$ is the $h$-value at the end of the previous trajectory, $h_{ij}^0$ the initial $h$-value of the edge, and $R$ scales the reinforcement of the edges. Confidence values are reinitialized at zero to start the next trajectory. This mechanism provides a built-in learning schedule such that the scale of the reinforcement signals grows progressively: rewards are initially small when trajectories are short, and they become larger when transitions are accurately captured in the model. During the deliberations, states are sampled from the prior ECM that did not receive the rewards yet, while the posterior ECM is updated with confidence and rewards at the end of the trajectory. It is equivalent to sampling states from the prior ECM, but updating the posterior ECM with the rewards $R$ for all edges in the trajectory every time a prediction was verified by the observation in the environment. Metaphorically speaking, this mechanism is analogous to layers of snow accumulating in time on salient features. At the end of a trajectory, the snow is cleared away, bringing all salient points back at an equal level.

To complete the update of the FEPS agent, the policy is modified according to the EFE inferred from the new world model and can be adjusted to promote an explorative or a conservative behavior. In particular, $h$-values of an edge $b_i \rightarrow a_j$ are set to the expected free energy in Equation \eqref{eq: EFE}
with a world model conditioned on $b_i$ and $a_j$. Each $h$-value directly carries the surprise expected from traversing the corresponding edge. As in \cite{Mazzaglia2022}, the policy is defined using a softmax function:
\begin{equation}\label{eq: policy from EFE}
\pi(a_j|b_i) = \text{softmax}(\zeta\  \mathcal{G}_{b_i}[a_j]).\end{equation}
where $\zeta$ is a (real-valued) scaling parameter and $\mathcal{G}_{b_i}[a_j]$ is the value of the EFE for action $a_j$ coming from state $b_i$. In active inference, $\zeta$ is typically negative. When it becomes more negative, actions associated with small EFE receive a large probability. More specifically, looking at the decomposition of the EFE in Equation \eqref{eq: EFE}, actions associated with larger entropies, that is lower certainty, together with higher chances of landing on preferable states or observations, become more attractive during the deliberation. In contrast, if the scaling parameter is positive, large EFE yield large probabilities in the policy, and actions with high certainty but also less chances of meeting preferences are more likely to be sampled. In this case, the agent implements a conservative policy that is confined to a known region of the environment at the expense of reaching the preferred states. When $\zeta = 0$, the policy is uniform, and the agent has no bias towards certainty nor utility. 

\section{Algorithmic features of the FEPS} \label{sec: FEPS algo features}

The FEPS can be augmented with a number of techniques that take advantage of the world model and its interpretability. During learning, the internal model can be leveraged to identify transitions that are instrumental to gain information or to get closer to a preferred state. Furthermore, the performance of an agent in completing a task can be enhanced by evaluating the correct belief state accurately and quickly. Since the policy depends on the EFE, the preference distribution can be tuned according to the task: to seek information to complete the model, or to complete a given goal. Therefore, we propose to separate training into two phases, depending on how the preference distribution is constructed. We introduce a belief state estimation scheme that distributes belief states over multiple clone clips and eliminates those that are incompatible with new observations.

\subsection{Leveraging preferences as a learning tool}

So far, the preference distribution entering the EFE was not defined. One can optionally leverage it to define a goal in the environment, be it for the purpose of gaining information, or to solve an actual task. Therefore, we propose to separate learning into two tasks: model the environment and attain a goal in it.
During the first phase, which we denote the wandering phase, the agent explores the environment without a prescribed goal. Instead, actions whose outcomes are expected to reduce prediction errors should be favorized. This phase spreads over multiple episodes and relies on interacting with the environment. In contrast to the wandering phase, the second phase is dedicated to learning to complete a given task. For this purpose, we designed an algorithm to infer a goal-oriented policy from the world model in a single step and without further information.

\subsubsection{Seek information gain about the world model}

Before designating any task bound to the environment as a preference, we investigate whether the preference distribution can be used to incentivize actions that minimize prediction errors, according to the current world model. This is directly related to the minimization of the VFE. Specifically, preferences should encourage the agent to seek transitions the world model associates with certainty $-$ or equivalently with high probabilities, irrespective of actions. Sequences of interaction steps with the environment guided by this preference distribution belong to the \textit{wandering phase}. To reflect the preference for highly probable transitions in the world model regardless of the action chosen, the preference distribution is constructed as the marginal of the world model over actions:\begin{align} \label{eq: pref wandering}
\text{pref}(B_{t+1}, S_{t+1}|B_t) 
&= \sum_{a} \pi(A_t=a|B_t)\, p(B_{t+1}|B_t, A_t=a)\, p(S_{t+1}|B_{t+1})\\
&= p(B_{t+1},S_{t+1}|B_t).
\end{align}
Plugging this distribution into the expected free energy evaluated for an action $a$ in Eq. \eqref{eq: EFE} results in the following:
\begin{align}
 \mathcal{G}_{b_t}[a_t] &= \sum_{b_{t+1},s_{t+1}} p(b_{t+1},s_{t+1}|b_t, a_t) \log \frac{p(b_{t+1}|b_t, a_t)}{\sum_{a} \pi(a|b_t) p(b_{t+1}|b_t, a) p(s_{t+1}|b_{t+1})} \\
&=\mathbb{D}_\mathrm{KL}\left[p(B_{t+1}|b_t, a_t) || p(B_{t+1}|b_t) \right]\\
&= \mathbf{IG}(B_{t+1}, A_t = a),
\label{eq: EFE wandering}\end{align}
where $\mathbf{IG}(X, Y=y) = \mathbb{D}_\mathrm{KL} [p(X|Y=y) || p(X)]$ is the information gain about the random variable $X$ when the value $y$ for the second random variable $Y$ is known. The dependency on the observations dropped from the first to second line thanks to the constraints the clone structure imposes on the emission function. A complete derivation of this formula is provided in Appendix \ref{appendix: EFE wandering IG}.

If we follow the conventional formulation of active inference as in section \ref{sec: FEP}, the agent should increase the probability of sampling an action that minimizes this EFE. Doing so during the wandering phase, the agent would therefore seek actions it estimates will yield the lowest information gain about belief states. As a result, the agent would stay in a region of the environment where it predicts it will receive the least surprise, according to its world model. This situation is sometimes referred to as the Dark Room problem \cite{Friston_2012_DarkRoom}: an agent that adapts by minimizing its surprise about observations would act to stay in a dark room instead of using actions to explore other places that may be more surprising, but also more favorable to its survival, because all observations there would be predictable. 

There is, however, an easy solution to this problem for FEPS. In order to avoid the dark room problem and to select actions that are expected to improve the model of the environment, the scaling parameter $\zeta$ in Eq. \eqref{eq: policy from EFE} can be set to a positive value. In this case, the larger the EFE associated to an action $-$ and the estimated information gain about the next belief state, the larger the probability of this action in the policy. The scaling parameter $\zeta$ can be understood as a way to determine how greedy an agent is in its exploration, or how strongly the information gain associated with each action influences its behavior. We call \textit{wandering phase} the interaction steps in which the agent samples its actions from such a policy.

\subsubsection{Task-oriented behavior by inferring preferences on belief states}

At the end of the wandering phase, a task is designated by encoding the associated targets with a high probability in the preference distribution. From there, the agent can plan, that is sample a sequence of actions to follow to achieve the goal. While the target is identified as a sensory state for the FEPS, transitions that are useful to reach it are deduced from the world model.  In our framework, updating the policy takes a single step and does not require further interaction with the environment.

Though active inference commonly determines the behavior by planning sequences of actions, it becomes expensive for distant horizons $T_h$. A sequence of $T_h$ actions must be chosen out of $N_\text{actions}^{T_h}$ possible combinations, by evaluating the EFE in a space of $(N_B \times N_S)^{T_h}$ possible outcomes for each sequence. Methods such as habitual tree search or sophisticated inference have been developed to mitigate this scaling issue \cite{Fountas2020, Friston_sophisticated_2021}. Alternative approaches are presented in \cite{Mazzaglia2022}. 

Instead of planning by evaluating the generative distribution over all possible future sequences of outcomes, we propose to encode the long-term value of a state directly into the preference distribution. Our scheme is reminiscent of successor representation \cite{Dayan1993, Russek2017, Momennejad2017}, in that it estimates a value function provided some expectations about occupancies of states in the future, either acquired by experience with reinforcement for example, or by inverting a learned transition function. In contrast to searching a tree of future sequences of actions, this method does not rely on mental time traveling \cite{Roberts2002}, to the extent that agents do not simulate possible future scenarios. Instead, they are ``stuck in time", and infer preferences in one go from stochastic quantities stored in the world model, in contrast to \cite{Nguyen2024_robustAIF}.  Our method also departs from sophisticated inference \cite{Friston_sophisticated_2021, Kawahara2022}: instead of bootstrapping expected free energies in time, we bootstrap preferences over belief states and calculate the EFE only once. 

We model the preference distribution to factorize over sensory and belief states \cite{Wei2024}, and we condition it on the current belief state $b_t$:
\begin{equation}\text{pref}(S_{t+1}=s,B_{t+1}=b|b_t) = \text{pref}(s)\ \text{pref}(b|b_t).\end{equation}
The first part, $\text{pref}(s)$, is an absolute preference distribution over sensory states, that is independent of where the agent believes it is in the world model. Since the environment is assumed to be partially observable, a target that is externally given can only be encoded in a shared state space, that is as a sensory state for the agent. More specifically, if $s^*$ is the target state for an observation, a probability $\text{pref}(S_{t+1} = s^*)=p^*$ is associated to it. All other observations are given a uniform distribution $\text{pref}(S_{t+1} \neq s^*)=(1-p^*) / (|\mathcal{S}|-1)$. The second part $\text{pref}(b|b_t)$ then reflects look-ahead preferences over belief states and reflect how useful an agent estimates a transition to be in order to satisfy its absolute preferences.  In other words, the utility of belief states over longer horizons is inferred from the value associated with the observations they can transition to. This way, even if the goal might be far away in the world model, the preference for the target observation propagates to intermediate belief states that contribute to reaching it. Metaphorically speaking, the preference for a target observation propagates to the belief states that are useful to get to that target observation. For example, consider an animal in a maze. The target is manifested with a high preference towards observing food. However, the preference distribution over the locations does not indicate how to reach the food. To remedy this, the agent infers the value of belief states in order to reach the target: if a transition to some location brings the animal closer to the food, it is assigned a higher probability in the preference distribution. This way, the preference distribution highlights the path of relevant actions to the target. 

We propose a heuristic algorithm to estimate the look-ahead preference distribution $\text{pref}(B_{t+1}|b_t)$ over transitions in the world model. The algorithm can update the policy $K_\text{pref}$ times if needed, to refine the preference distribution. The initial policy $\pi^{(0)}(A_t|B_t)$ is uniform. During each update $k$, two quantities are calculated: (1) the look-ahead preference distribution results from the value of each transition, that estimates how useful a transition is to reach a target within a prediction horizon $T_\mathrm{h}$, and (2) the policy $\pi^{(k+1)}$ is calculated with Equation \eqref{eq: policy from EFE} and the latest preference distribution. 

To initialize each update step $k$, the policy $\pi^{(k)}$ is used together with the world model to evaluate how easy a belief state can be reached from the current state, in a distribution we denote \textit{reachability}:
\begin{equation}
r^{(k)}(b_{t+1}|b_t) = \sum_a p(b_{t+1}|b_t,a) \pi^{(k)}(a|b_t).
\end{equation}
The reachability of $b_{t+1}$ coming from $b_t$ is large if there exists transitions $b_t \xrightarrow{a} b_{t+1}$ associated with high probabilities in the world model and the corresponding actions have high chances to be sampled in the policy.

In addition, the initial value of a belief state equals the value of the observation it corresponds to:
\begin{equation}
    v_{ 0}(b_{t}) = \sum_{s}\text{pref}({s})\, p({s}|b_{t}).
\end{equation}
For the clone-structured model, this sum reduces to a single term. At this stage, the only belief states associated with a high value are those that represent a target observation in the absolute preference distribution. 

Next, for each iteration $n \in [1, T_h]$, where $T_h$ is a prediction horizon, the value of a belief state is increased if it can lead to transitions that are useful to reach the target within $n$ steps in the environment:
\begin{equation}
v_{{n}}(b_{t+1}) = \max \left\{v_{{n}-1}(b_{t+1}), \ \max_{b^+} \{\beta^{{n}-1}\, r^{(k)}(b^+|b_{t+1})\,v_{n-1}(b^+) \} \right\},
\end{equation}
for $0\leq \beta \leq 1$ some discount factor that makes the value of the state decrease with the number of steps between this state and the target. At each ${n}$, the value of a belief state $b_{t+1}$ can either keep its previous value $v_{{n}-1}(b_{t+1})$, or it can take the discounted value of the best state $b^+$ it can transition to. As a result, the value of a state can only increase. If ${n}$=1, using this value function can point at the right decision while being one step away from $s^*$. However, it does not incentivize the correct action when starting further away from the goal.  To mitigate this effect, a state can inherit the value of the states it can reach over larger time scales ${n} > 1$, thereby propagating the preference for the target to more distant but useful belief state states.

When the prediction horizon is reached, a transition $b_t \rightarrow b_{t+1}$ is associated with the following probability in the look-ahead preference distribution:
\begin{equation}
\text{pref}^{(k)}(b_{t+1}|b_t) = \delta_{b_{t+1} \in \mathrm{ch}(b_t)}\ \frac{v_{n=T_h}(b_{t+1})}{\sum_{b'_{t+1}\in\mathrm{ch}(b_t)} v_{n=T_h}(b'_{t+1})}
\end{equation}

where the set $ch(b_t) = \{b^+\ |\  r^{(k)}(b^+|b_t) > \overline{r^{(k)}(B_{t+1}|b_t)}\}$ contains the children of $b_t$, or the states that are easily reachable from $b_t$. $\overline{r^{(k)}(B_{t+1}|b_t)}$ is the mean reachability of belief states coming from state $b_t$.

Finally, to conclude the $k$-th update step in the algorithm, the policy $\pi^{(k+1)}$ is calculated by using the preference distribution $\text{pref}(S_{t+1})\,\text{pref}^{(k)}(B_{t+1}|B_t)$ in the expected free energy, as in Eq. \eqref{eq: policy from EFE}. We show a possible look-ahead preference distribution in the world model in figure \ref{fig: belief in superposition}.

\subsection{Delineate belief states for the same observation}

\begin{figure*}[t!]
    \centering
    \includegraphics[width=0.75\linewidth]{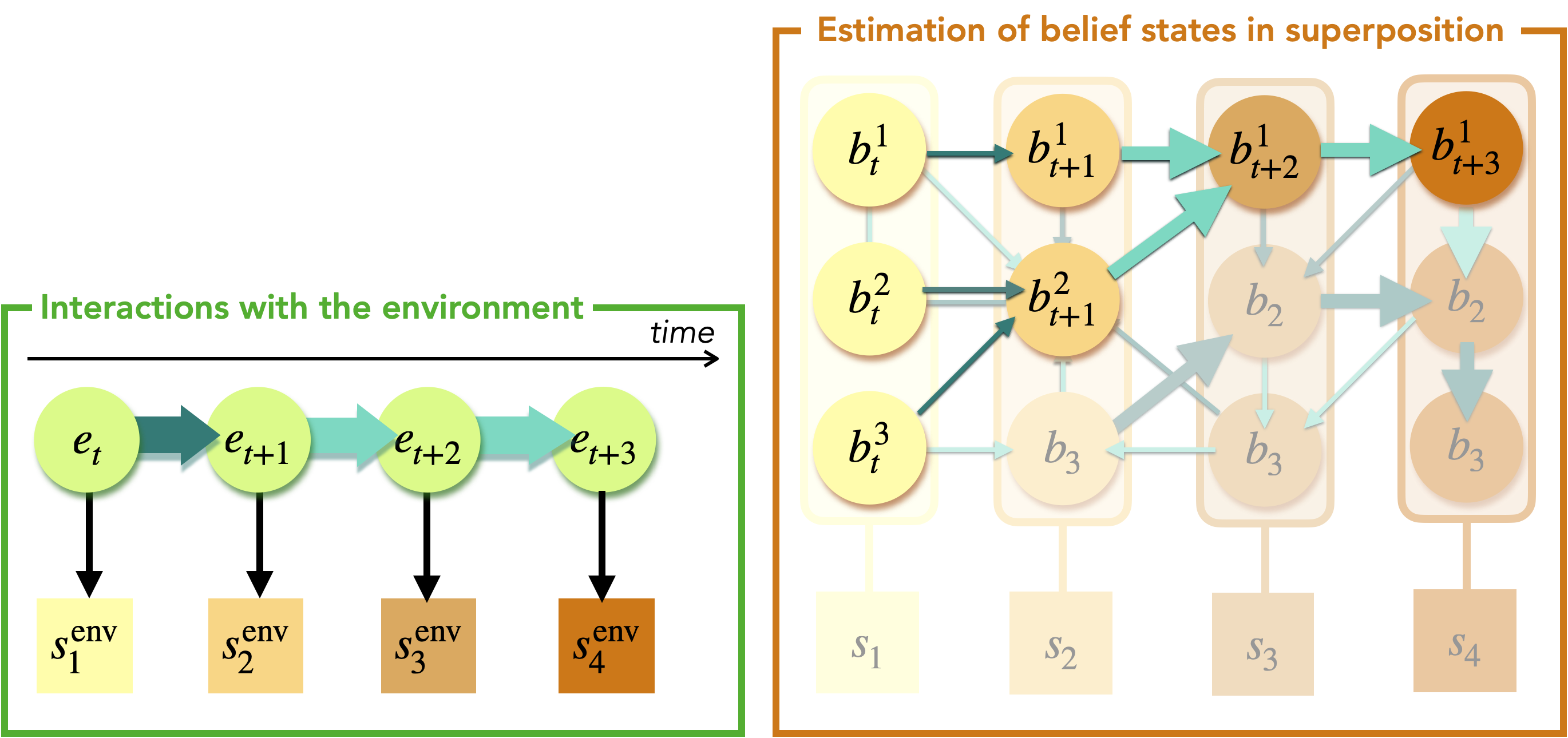}
    \caption{\textbf{Estimation of belief states in superposition, after the world model has been trained.} To minimize its prediction error due to faulty belief state estimation, an agent considers  multiple clone clips as candidate belief states simultaneously. For the initial observation, $s_1^\mathrm{env}$ (on the left), the agent includes all corresponding clone clips $\{b_t^i\}_{i=1}^3$ to its hypothesis, as depicted on the right. Conditioned on the chosen action, a clone clip is sampled for each candidate belief state to represent the next one. Finally, all clone clips that are incompatible with the observation from the environment are eliminated from the hypothesis. The clips that remain become the current candidate belief states. In the world model, the thickness of the arrows represents the look-ahead preferences: the larger the arrow, the more advantageous is the transition in order to reach the target observation, $s_4$ in this case.}
    \label{fig: belief in superposition}
\end{figure*}

In spite of prescribing a method to sample actions, active inference does not include techniques to efficiently choose belief states when multiple of them could explain the current observation. In particular, models involving neural networks lack the interpretability to design suitable belief state selection methods.
Therefore, letting the agent learn a world model with a clone-structured HMM has advantages beyond planning. We propose a technique to evaluate belief states \textit{in superposition}, depicted in Fig. \ref{fig: belief in superposition}. When placed in an environment and receiving its first observation, the agent makes an initial hypothesis about its belief state by distributing an excitation to any clone clip compatible with the observation, as shown in Figure \ref{fig:FEPS model}c. At each step, an action is sampled from the policy for each excited clone clip. The resulting frequencies define a new distribution, from which an action is sampled before it is applied to the environment. For each compatible clone clip, the agent samples a new clip to represent the belief state it anticipates it would transition to if the clone clip under consideration stands for the correct belief state. The excitation jumps onto the new clip. After applying the action to the environment and receiving the resulting sensory signal, the agent takes away the excitation on any clip that does not match the current observation. Depending on the structure of the environment and the number of clones for each observation, the agent progressively narrows down its candidate belief states to a single clone clip, in spite of the initial uncertainty. In the event that the world model is imperfect and the agent has eliminated excitations on all clips, it starts its hypothesis over, and considers all the clone clips of its current, unpredicted observation as candidate belief states. This mechanism allows the agent to evolve in environments with ambiguous observations in the absence of more contextual information about its initial conditions. 

\section{Numerical results} \label{sec: numerical results}

In this section, we present a numerical analysis of the model on environments inspired from behavioral psychology tasks, namely a timed response experiment in a Skinner box and a navigation task to forage for food. The parameters used for the simulations are given in Table \ref{tab: simulation parameters} in the appendix.

\subsection{The timed response task} \label{subsec: Skinner box}

\begin{figure*}[t] 
    \centering
    \includegraphics[width=0.75\linewidth]{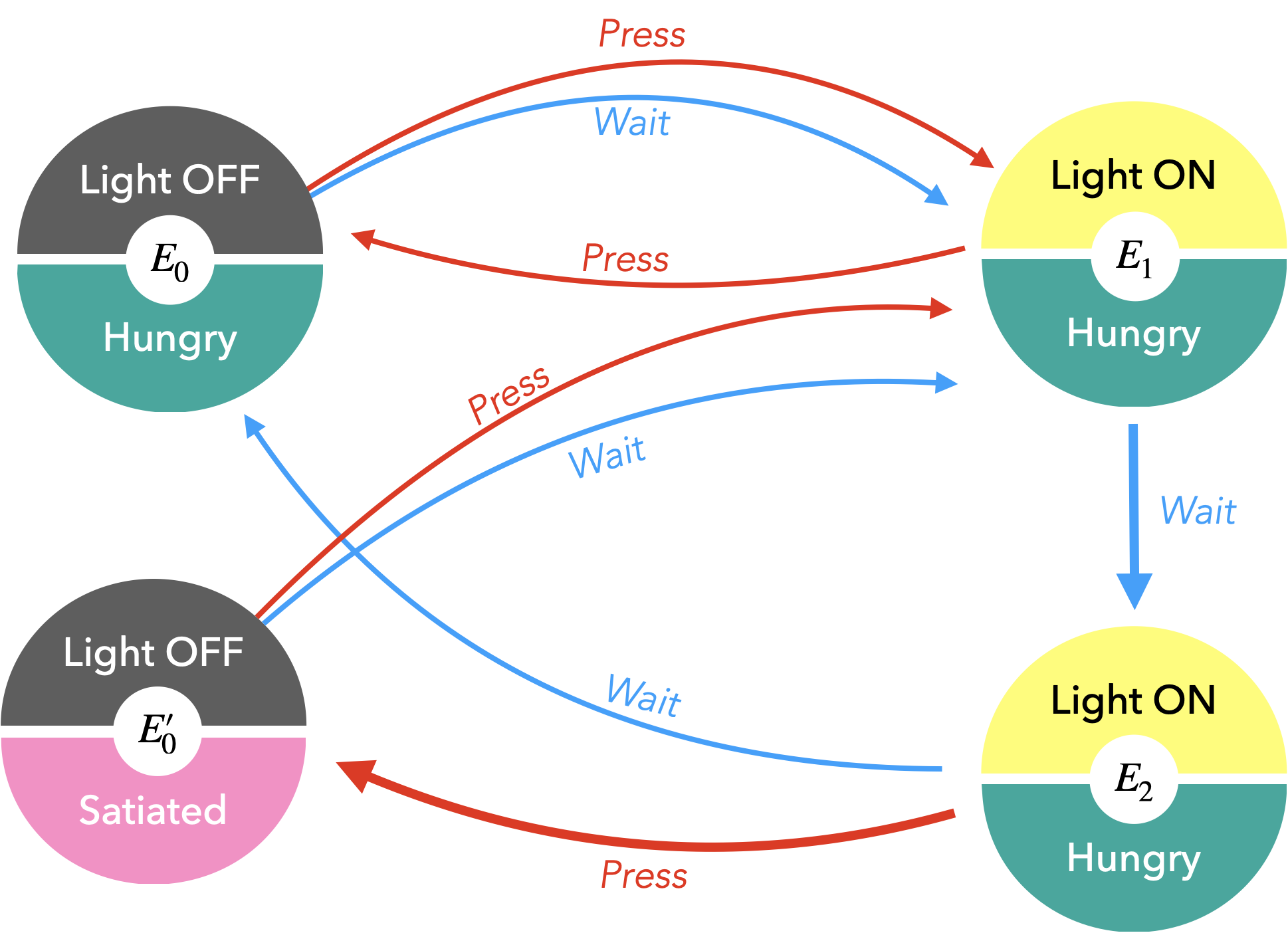}
    \caption{\textbf{MDP for the timed response environment} This environment has four hidden states. The observations are compositional and contain information that are both external (light on or off) and internal (hungry or satiated) to the agent. Arrows correspond to the transitions the actions the agent can result in. In this environment, the agent can either wait or press a lever.  In order to complete the task, the agent must reach $E_0'$ and feel satiated. The only way to do so is to follow the actions marked with thicker arrows. The observation (light on, hungry) is called ambiguous because it can be emitted by two hidden states $E_1$ and $E_2$ that can only be distinguished with context.}
    \label{fig: MDP Skinner Box}
\end{figure*}

\subsubsection{Learn short-term associations}

The timed response task is a minimal environment for an agent to learn to contextualize its observations with past states and actions when the sensory signals emitted by the environment are ambiguous. This environment simulates an animal standing in front of a door that can be opened with a lever. The goal is for the agent to learn a conditioned response and to press a lever at the right time in order to access food. The environment's MDP is depicted in figure \ref{fig: MDP Skinner Box}. For this task, the observations combine two sensory inputs: $\mathcal{S} = $ \{ (light off, hungry), (light off, satiated), (light on, hungry)\}. Since food can be consumed only when the light is off, the observation (light on, satiated) in excluded from the set. The actions are $\mathcal{A} =$ \{wait, press the lever\}. The environment is initialized in the (hidden) state $E_0$, that emits observation (light off, hungry). From there, the light turns on, regardless of the action taken by the agent. Once the light has turned on, the agent must learn to wait one step before pressing the lever. If it does so too early, it gets back to the initial state. If the agent activates the lever on time, the environment transitions to state $E_0'$, where the light is off, but the agent is satiated. 

For the simulations, we give each observation two clones. It makes it possible to accommodate enough candidate belief states to model up to two ambiguous hidden states that emit the same sensory signal, enabling the agent to adapt its policy to a one-step waiting time between the light turning on and the food being accessible for example. When observations are not ambiguous and can be emitted only by a single hidden state in the environment, some belief states become redundant. This redundancy can make the training more challenging to the extent that the agent has to find a convention and adapt its model accordingly before it can account for all transitions in the environment faithfully. The larger the waiting time $n$, the more clones might be necessary to learn. We train 100 agents for 4000 episodes of 80 steps in this environment and we test two scenarios. In the first, the agent is directly given the preference for the target observation (light off, satiated). The EFE is then scaled with a scaling parameter of $\zeta = -1$ in the policy in Eq. \eqref{eq: policy from EFE}. In the second scenario, we test whether the agent can learn more efficiently if it wanders aimlessly through the environment for the same number of episodes without preference for the target before adapting its policy to the task, with a scaling parameter of 0.

\subsubsection{Simulation results}

\begin{figure*}[t!] 
    \centering
    \includegraphics[width=1\linewidth]{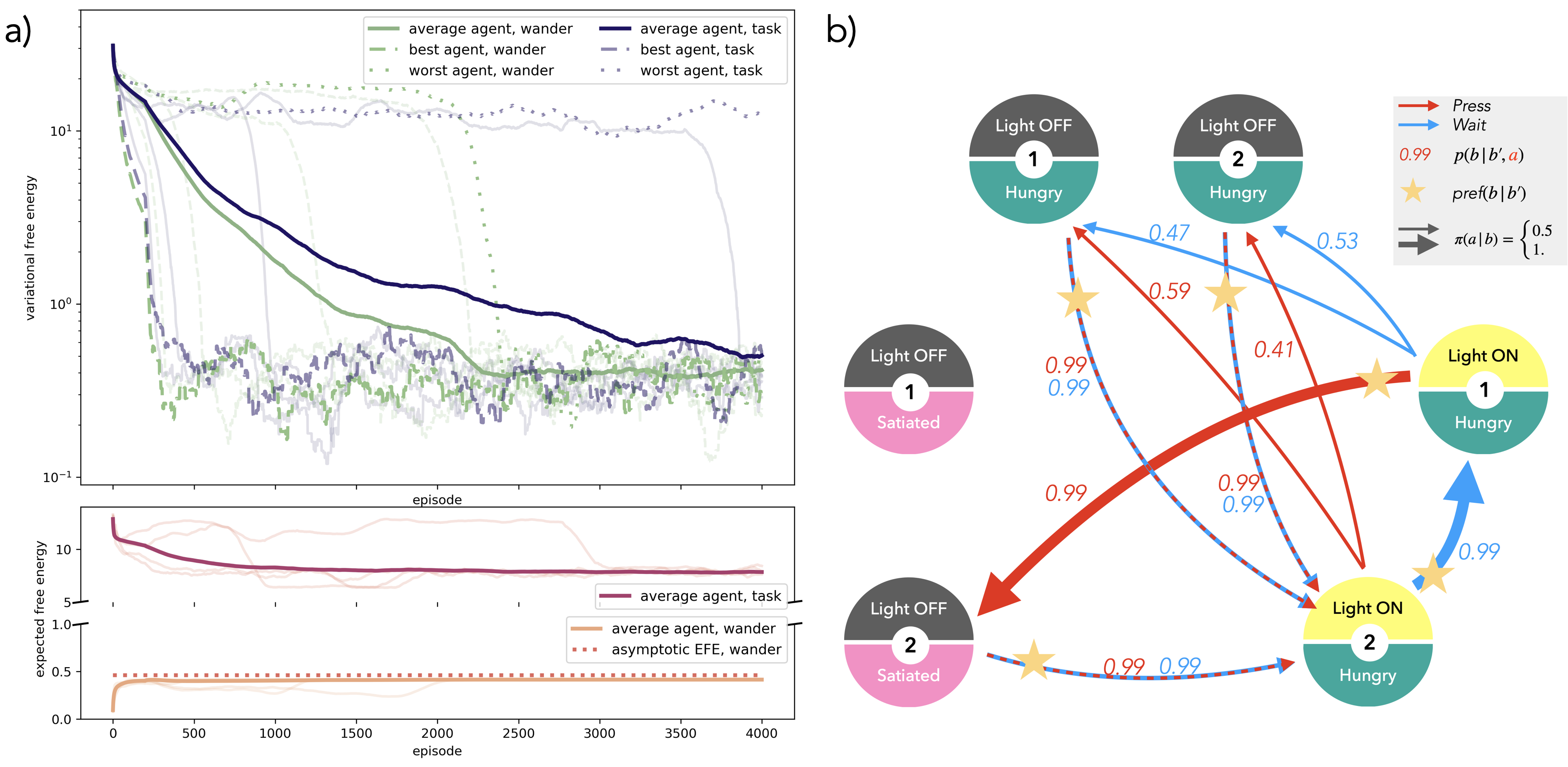}
    \caption{\textbf{Training FEPS agents for the timed response task} a) Evolution of the variational free energy (top) defined in Eq.\eqref{eq: VFE} and expected free energy as in Eq. \ref{eq: EFE} (bottom) during the training, averaged over 100 agents and a time window of 100 episodes. At each step, the VFE depends on the specific belief states and actions that were sampled. Two types of training are compared: a first set, ``task" in dark purple, learned the model with a preference to fulfill a task in the environment, while the second set, ``wander", in green, experimented aimlessly in the environment with a uniform policy before switching to the task. Both were trained for 4000 steps before being tested on the task. The best and worst agents are represented with dashed and dotted lines respectively, and examples of individual agents were traced with transparent lines, full for task-oriented agents, and dashed for the wandering ones. When the VFE converges to its minimal value, the world model is precise enough for most belief states to make planning possible. As expected from the values chosen for the scaling parameter $\zeta$, agents select actions that minimize and maximize the EFE for task-oriented and wandering agents respectively. The EFE of wandering agents plateaus quickly at the limit derived for Appendix \ref{appendix: limit EFE wandering}.   b) World model learned by one of the agents trained on the task, where each circle is a belief state, whose observation is denoted by its colors and label. The numbers at the center of the circles are the clone indices for each clone clip. Arrows indicate the transitions learned in the world model, red for action ``press the lever", and blue for ``waiting". Dashed lines indicate that both actions lead to the same transition. The weight on the arrow indicates its probability in the world model. Stars mark transitions that were identified as useful to achieve the goal with a probability of 1 in the preference distribution. The policy is indicated by the thickness of the arrows, where a thick arrow corresponds to probabilities close to 1, and thinner close to 0.5. }
    \label{fig: DR}
\end{figure*}

The timed response environment is a minimal testbed for the FEPS, where some hidden states are not uniquely defined by the observation they emit, but also by the recent past. The agent must learn two types of belief states. While clones for (light off, hungry) and (light off, satiated) only support transitions that are independent of the actions the agent takes, the clones for (light on, hungry) must appropriately use information about the previous observation and action to be contextualized and distinguished. Some results are reported in figure \ref{fig: DR}. 

During training, regardless of the strategy that the agent uses to resolve the environment, all agents follow a similar learning pattern. The acquisition of the world model happens in stages, where the transition from one to the next is manifested in a steep increase in the length of the trajectories of correct predictions, or equivalently, as a steep decrease in the free energy, as in figure \ref{fig: DR}a. First, agents quickly eliminate transitions that are impossible in the environment, leading to an initial drop in the VFE. For example, as in figure \ref{fig: MDP Skinner Box}, a direct transition from (light off, hungry) to (light off, satiated) is incompatible with the timed response task. During the second phase, the number of rewards the agent can collect is limited by the absence of convention on the context-sensitive belief states. Therefore, a plateau is observed on the VFE. A final sharp decrease in the VFE signals the adoption of a convention between clones to accurately disentangle ambiguous hidden states that cannot be told apart with observations alone. The EFE evolves as expected during the training: it decreases for the task-oriented agents, while it rapidly plateaus to its limit (see Appendix \ref{appendix: limit EFE wandering}) in with a wandering phase. However, in contrast to the VFE, the convergence of the EFE to its asymptote value does not reflect the fact that the model is good enough to make accurate predictions.

For most agents, training with or without a preference for the target in order to construct the world model does not influence the final model. As in figure \ref{fig: DR}a, the two best agents, that is, those that converge to free energy values below 1 the earliest, share similar learning curves. The averaged behaviors remain fairly identical, except for the length of the second stage of the learning phase, where the agents have yet to adopt a successful convention. In particular, in the absence of aimless experimentation with the environment, this second phase can last longer, such that convergence is not attained for a fraction of the agents trained in this way.

After convergence of the free energy, or equivalently of the number of successful predictions, the model has collapsed on a single representation of the environment, where the ambiguous observation is contextualized by the previous belief state and action. For example, the observation (light on, hungry) is divided into two belief states with different uses, reflecting the two hidden states in the environment. The corresponding clone clips bear more information than the observation to which they are linked. In figure \ref{fig: DR}b, clone 2 for (light on, hungry) is the first hidden state encountered when the light turns on, and clone 1 can only be accessed by waiting from clone 2. When more clones are provided than necessary, two possibilities can arise: either the agent uses all the clones as duplicates of the same hidden state in the environment (as for clones of (light off, hungry)), or a single clone is reachable (clone 2 of (light off, satiated)) and the other is never trained because it was not visited. When multiple clone clips participate equally in the representation of the observation, the actions that lead to it also split with equal probability (as derived in Appendix \ref{appendix: limit EFE wandering}).

Finally, for the agents that converged to an appropriate representation of the environment, the look-ahead preferences inferred from the world model result in an optimal policy when using the EFE. A visualization of the preferences and policy is provided in figure \ref{fig: DR}. We chose a prediction horizon of 2 steps, and a single iteration was required to calculate appropriate look-ahead preferences for the transitions. The transitions with maximal preference were kept so that, coming from a belief state $b_t$, a single belief state $b^*$ is more preferable than the others. The agents were tested for 1000 rounds, each starting in $E_0$ in the environment. For a prediction horizon of a single step, the edge between clone 1 of (light on, hungry) and clone 2 of (light off, satiated) would be the only edge to carry a probability larger than other transitions in the preference distribution. Thanks to the look-ahead preferences (and the adoption of a convention between belief states), waiting between clones 1 and 2 of (light on, hungry) is also preferred. The policy that results from this preference distribution is optimal to solve the task, in spite of initially having no hints about the target prior to the last transition.

\subsection{Navigation task in a partially observable grid}

\begin{figure}[t!] 
    \centering
    \includegraphics[width=1\linewidth]{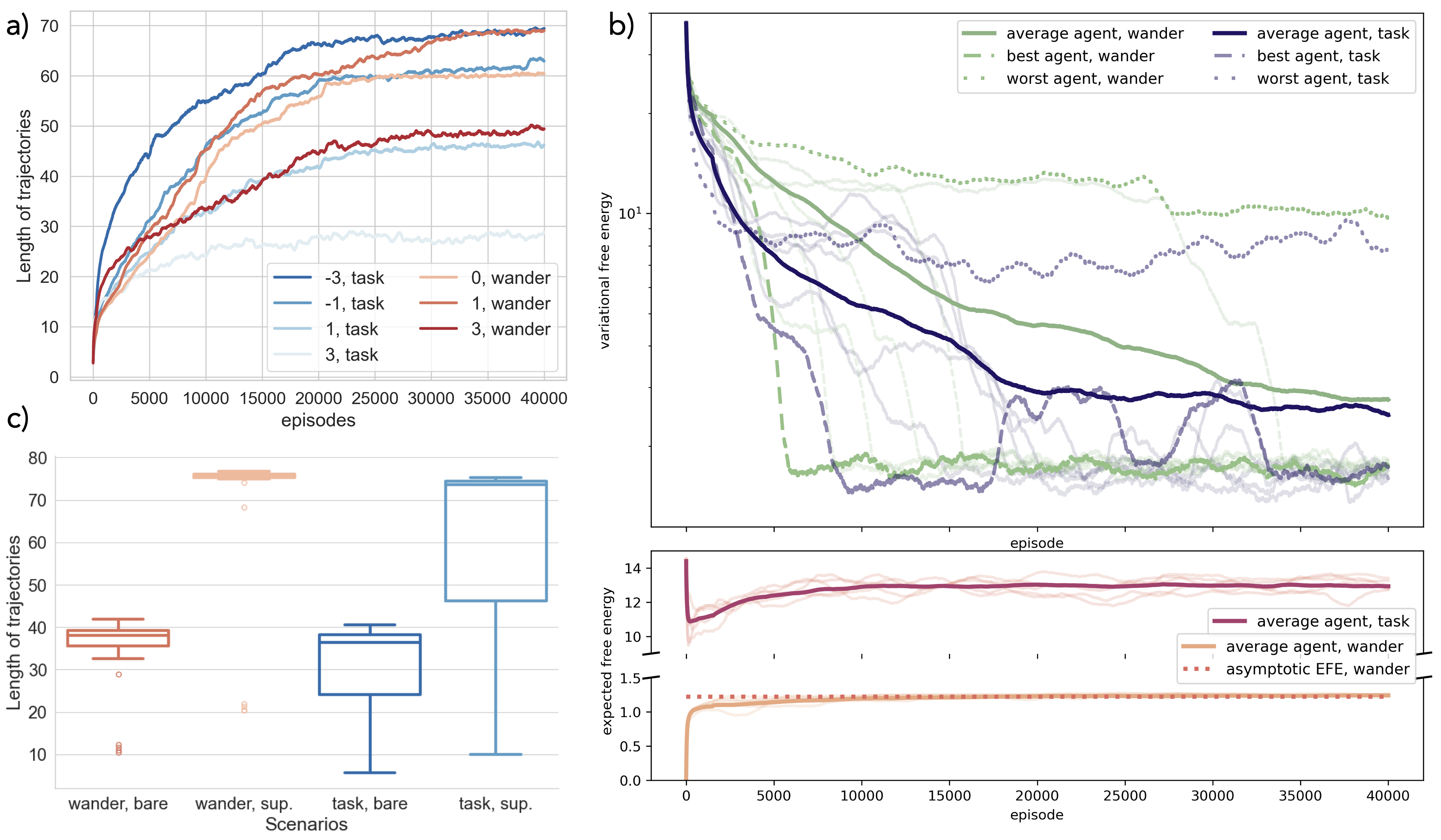}
    \caption{\textbf{Training results for the grid world environment.} a) Evolution of the length of the trajectories during the training, for different scaling parameters ranging from -3 to 3, and different preference distributions: the agent can either learn to complete the task from the start (``task"), or first wander in the grid (``wander").  We represent the running averages over a time window of 1500 steps of the lengths of trajectories, averaged over 30 agents. These lengths depend on the specific belief states and actions sampled by the agents. 
    b) Evolution of the variational and expected free energies during the training for the two best settings in a): the task-oriented preferences are paired with a scaling parameter of -3, and the wandering preferences with a parameter of +1. The thick lines represent the running average of energies over a time window of 1500 steps, averaged over all 30 agents, while the dotted and dashed lines stand for the best and worst agents, that is the agents whose VFE converge first and last, respectively, to a minimum. The transparent lines indicate the behavior of agents selected randomly: these lines are full for task-oriented agents and dashed for wandering agents.
    c) Comparison of the accuracy of the model to make predictions between the two best parameter settings in a). The policies of the agents are uniform,  such that the actions yielding most certain outcomes cannot be relied upon to validate predictions. Two belief state estimation techniques are tested. Bare estimation samples a single clone clip at a time, whereas the evaluation of belief states in superposition allows multiple clone clips to represent candidate belief states simultaneously, as long as they produce predictions that are compatible with the next observation. Each individual agent is tested over 1000 trajectories of at most 80 steps.}
    \label{fig: GW}
\end{figure}

\subsubsection{Long-term planning in an ambiguous environment with symmetry}
The FEPS is further challenged in a larger navigation task, where observations are shared among hidden states, and multiple sequences of actions can emit the same observations, due to the symmetry of the environment. In order to disentangle the hidden states, the agent must use long-term information about its past observations and actions to contextualize its current state in a way that is consistent across actions. In this environment, food is hidden in one position in a grid. Locations in a 3 by 3 grid world are associated with smell intensities, denoted by their integer intensities from 0 to 3 from the lower left to the upper right corners, according to their closeness to the food, and represented by increasingly warm colors in figure \ref{fig: GW test}a. 

The world model is provided with 3 clones for each of the four observations, and the behavior repertoire comprises directional actions $\mathcal{A} = $ \{go right, go left, go up, go down\}. 30 agents were trained on different hyperparameters and preferences, for a total of 40000 episodes of 80 steps. The hyperparameters are provided in the appendix. Since this environment is larger and more complex, we test different training techniques, by changing the preference distributions and varying the scaling parameter. In particular, we test two scenarios. In the first the agent is trained with preferences pointing at a target in the environment, while in the second, the preferences are identified with the marginal of the world model over actions, which incentivizes dissociating the effect of actions on the environment. While in the first case, the agent tries to directly solve the task, in the second, the agent wanders first before it learns the task. For each scenario, different scaling parameters were implemented, ranging from -3 to 3 for the task-oriented training, and from 0 to 3 with a wandering phase. Note that choosing a scaling parameter of 0 while training either directly on the task or with a preceding wandering phase results in the same uniform policy.

\subsubsection{Simulation results}

\textbf{Training task-oriented and wandering agents.} Compared to agents trained with a uniform policy, agents using policies inferred from the EFE can achieve longer trajectories. This is apparent in figure \ref{fig: GW}a), where with a scaling parameter of 0, the trajectories of agents reach a length of 60 on average by the end of the training, whereas given a proper tuning, agents using the EFE to define the policy can predict up to 69 transitions, regardless of the training method. Adapting the scaling parameter to the preference distribution is crucial for the agents to learn the model and query actions that result in longer trajectories. For task-oriented agents, as expected from Equation \eqref{eq: EFE}, a negative scaling parameter at -3 is optimal, whereas +1 yields the best results for the wandering agents. For both training methods, setting it at +3 results in the shortest trajectories, that are 28 and 50 steps long, respectively. In the case of task-oriented agents, a possible explanation is that in this regime and when the preference distribution designates a target in the environment, the policy that is derived from the EFE in Eq. \eqref{eq: EFE} minimizes the drive of the agent to explore and to fulfill any preferences in the environment. Therefore, the agent has no incentive to go outside of a region of the environment where it can predict its observations, and does not try to learn the rest of it. In contrast, a scaling parameter $\zeta = +3$ makes wandering agents very greedy. It is then possible that previously explored regions are not visited often enough to reinforce the correct associations in the world model in the long-run. Finally, our simulation shows that for the best parameter settings, the task-oriented agents converge on average faster than the wandering ones: while 17500 episodes are needed to predict 65 steps for the former, the latter crosses this milestone at 28000 episodes.

Looking at the evolution of the free energy during the training in figure \ref{fig: GW}b), one sees that in both cases, the VFE decreases as the length of trajectories increases. It looks as if agents minimize their free energy by maintaining a world model and improving their predictions about future sensory states. This matches the free energy principle. Comparing individual learning curves, wandering agents have more diverse behaviors than those trained on the task. We define the best and worst agents as those converging the earliest and the latest to a free energy minimum, respectively. The best wandering agents can converge faster than the task-oriented ones, whereas the worst wandering agent lands on a higher free energy by the end of the training. However, the free energy of the worst wandering agent either goes down or plateaus, but it does not rise again, as for the worst agent directly trained on the task. When the training scenarios influence the scale of the EFE, no significant variations around the average behavior is observed in the EFE to distinguish successful agents from those that did not converge. In particular, the EFE wandering agents converge very quickly to its asymptotic value, regardless of the accuracy of the world model.

In order to fairly compare the models obtained from each training method, note that whenever the length of trajectories or the free energy are optimized, two contributions are at play. On the one hand, training the world model to capture the environment more accurately decreases the uncertainty about the transitions following actions, and therefore improves the length of the trajectories and the free energy. But on the other hand, as long as the policy deviates from the uniform distribution, it influences how much uncertainty the agents seek. Therefore, in order to evaluate how much the world model helps making accurate predictions, we test the length of trajectories with a uniform policy for the agents trained in the two best parameter and preference settings. The trajectory lengths averaged over all agents trained with the respective parameters are provided in figure \ref{fig: GW}c). 

In spite of enabling faster convergence with less deviations in the final free energies, learning with task-oriented preferences deteriorates the prediction abilities to some extent compared to wandering first. Indeed, when tested with a bare belief estimation technique, where the agent can entertain a single belief state at a time, the median length of trajectories is slightly higher for the wandering agents, and fewer deviations are observed between these agents: while 50\% of the wandering agents can predict between 36 and 40 steps without mistake, this range expands to the interval between 25 and 39 steps for the task-oriented agents. Best and worst behaviors remain similar between both training techniques.

Estimating belief states in superposition doubles the length of trajectories for both training settings, with a stronger improvement for wandering agents. Indeed, the length of trajectories is increased to 76 steps for all wandering agents whose model converged. While the length of trajectories is doubled for all task-oriented agents, large discrepancies are noticeable by the spread of the inter-quartile space ranging from 47 to 73 steps. We expect that if the task-oriented agents were trained with 0 as a target observation instead of 3, the ambiguity of the lower left corner of the grid would not be lifted, and the length of the trajectories would decrease. This would also affect the ability of the agents to adapt to different goals in the environment.\\

\begin{figure*}[!t] 
    \centering
    \includegraphics[width=1\linewidth]{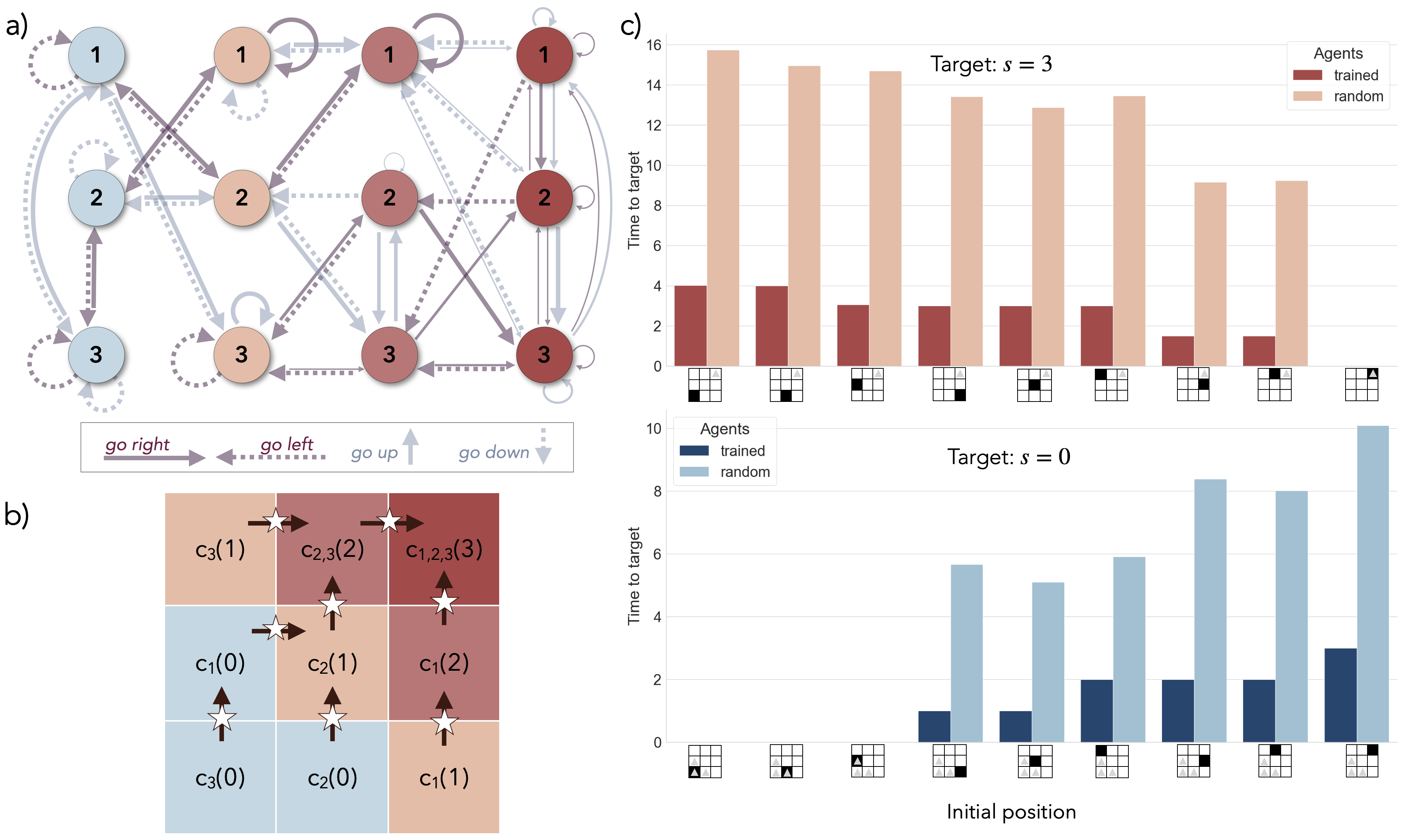}
    \caption{\textbf{Model and robustness to reward reevaluation for the grid world environment.} 
    a) Example of a world model learned by an agent directly trained on the task, with the target positioned in the top right corner of the grid. The circles represent the belief states as in fig.\ref{fig:FEPS model}, numbered with clone indices, and colored according to the observation they relate to in the grid. The arrows stand for the transition probabilities: the thicker the arrow, the more the agent believes taking an action from a belief state will lead to the state the arrow points at. 
    b) Mapping of the world model in a) onto the grid: each clone is associated with exactly one cell, that is, a single hidden state in the environment. Stars stand for the preferred transitions, and the arrows for the policy resulting from these preferences. 
    c) Median number of steps from each initial position in the grid in order to reach the target, where belief states are estimated in superposition. Two targets are given to the agent and symbolized with a gray triangle: reach observation of 3 (top right, red) and 0 (bottom left, blue), each requiring opposite policies. When targets were swapped, no interaction with the environment was necessary to re-evaluate the value of the transitions and the resulting policy. The median time to target is compared to that of a random agent with uniform policy and tested with the same procedure.}
    \label{fig: GW test}
\end{figure*}

\textbf{Exploit the world model to complete tasks flexibly.} After the free energy converges, the belief states in the world model no longer stand only for the observation they relate to, but also convey information about the path they can appear in. As an example, the world model of an agent trained directly on the task is represented in figure \ref{fig: GW test}a). Because the environment the agent was trained in is deterministic, actions tend to map onto a single belief state when the number of clones for an observation matches the number of hidden states in the environment. 

In particular, in spite of being agnostic of the nature of the environment, the world model can be interpreted as a topological map of the grid, as pictured in figure \ref{fig: GW test}b). Each clone maps onto a single location of the grid, and this mapping is consistent for all actions. When too many clones were provided compared to the degeneracy of an observation, multiple clones are associated with a single location, as for observation 3 in the upper right corner of the grid. This consistent mapping suggests that the agent is able to contextualize the observations with a sequence of previous states and actions. Indeed, before adopting a convention, starting from the bottom left corner, the agent cannot distinguish between going up or going right based only on its observations, because of their symmetric distribution in the grid. Instead, using long-term context, the two paths can be identified and carried over by different clones. Thanks to this mapping of clones onto single locations, the look-ahead preferences over belief states encode paths ending at the target. They steer the policy toward an optimal one.

Provided the target observation is signaled in the preference distribution, the agents can adapt their policy to reach it following near optimal trajectories. In figure \ref{fig: GW test}c), agents whose model successfully converged are tested to reach targets with observations 3 (top, red) and 0 (bottom, blue), requiring opposite policies to be attained. For both tasks, irrespective of the initial location of the agent, the performance is drastically improved compared to a random agent with uniform policy. In order to move the target, no other interaction with the environment was required than changing the preference distribution over sensory states, and no new observation was exchanged. 

While the paths towards observation 0 are mostly optimal, there is an overhead by at most one step to reach observation 3. Since the policy is an optimal one, as shown in figure \ref{fig: GW test}b), we suggest that this is due to the estimation of the belief states. In particular, the small size of the grid prevents agents from narrowing down their hypothesis to a single belief state early enough to opt for the right policy at the boundary of the grid. This could explain why on average, agents take 1.5 steps to reach the target when being right under it: they have 0.5 probability of finding the wrong belief state, and therefore choose the wrong action 50\% of the time. This eliminates the last erroneous state from the hypothesis and the agent behaves optimally in the next step.

\section{Discussion and outlook} \label{sec: Discussion}

In this work, we propose an interpretable, minimal model for cognitive processes that combines associative learning and active inference to adapt to any environment, without any prior knowledge about it and independently of any goal tied to it. We develop the Free Energy Projective Simulation model, a physics-inspired model that seeks inspiration in current paradigms in cognitive sciences, namely active inference, that encompasses the Bayesian brain hypothesis (BBH) and fits more broadly into predictive processing. A FEPS agent is equipped with a memory structured as a graph, the ECM, where the agent encodes associations between events represented as clone clips. As in models of associative learning in the cerebellum \cite{Thompson1998, Wagner2021, Rao1999},  internal reinforcement signals based on the prediction accuracy of the agent reinforce associations between observations. Clones clips are the support for belief states and acquire contextual meaning as the agent collects longer sequences of correct predictions about its observations and improves its world model. The behavior of FEPS agents does not depend on any reinforcement and is fully determined with active inference, such that the policy optimizes the expected free energy estimated for each action, given the current world model. We perform numerical analysis of the agents in two environments inspired from behavioral psychology: a timed response environment, as well as a navigation task in an partially observable grid world.

Leveraging the interpretability of the model, we identified three capacities a FEPS agent requires in order to interact with an environment and to reach a goal while controlling it, that is, minimizing its surprise about future events. (1) The representation of the environment in the world model should be accurate enough in its predictions, i.e. it should be capable of predicting future sensory signals. (2) The agent should be able to choose the appropriate belief state for aliased observations. (3) For a given time horizon, the agent should be equipped with an efficient mechanism to plan its course of actions.

First, in order to build a complete and accurate representation of the environment, we propose to start learning with a wandering phase. It focuses on exploring the environment strategically rather than completing a task. For this, we provide a model-dependent preference distribution that evolves as the agent learns and we show that the corresponding expected free energy of an action is equal to the information gain about belief states resulting from this action. As a result, the behavior of the agent is motivated by the acquisition of information in its world model rather than by a target bound to the environment. Our numerical simulations showed that a model learned in such a way is able to predict longer sequences of observations when actions are selected at random, compared to the model trained directly on a task. Previous works usually involve either adopting a different definition for the expected free energy \cite{Millidge2020}, or more commonly, adding terms to it to encourage explorative behaviors, \cite{Nguyen2024_robustAIF}, leveraging existing literature on curiosity mechanisms, for example \cite{Tinker2024, Oudeyer2007, Kim2019_curiosity-bottleneck, Pathak2017}. Alternatively, a count-based boredom mechanism \cite{Bellemare2016} could be well-suited for FEPS, as it could be directly implemented on the edge attributes, such as the confidence, to deviate the agent from transitions it has already resolved.

Second, we design a simple procedure to progressively disambiguate belief states with sequences of observations. It can be used to interact with the environment after the model has been trained. Leveraging the clone structure imposed to the clips in the memory of the agent, belief states can be estimated in superposition. As predictions for each candidate belief state are validated or not by the observation collected from the environment, the agent can eliminate candidate belief states that are incompatible with the context provided by its current sensory state. We show in our numerical analysis that this simple mechanism allows to double the number of correct, consecutive predictions, regardless of the technique used to learn the world model.

Thirdly, we introduce a planning method based on the expected free energy. Instead of relying on a tree search or a simulation of future scenarios, we propose to encode the utility of a transition between two belief states in the preference distribution. For this, we factorize the preference distribution into an absolute preference distribution that designates a target among the possible sensory states, and a look-ahead preference distribution. The latter assigns a probability to transitions between belief states that is commensurate with its estimated utility in reaching the target within a given number of actions. Using the world model, the value for each transition is determined in an iterative manner. We tested this scheme in two numerical experiments and achieved optimal policies in both cases, as long as the world model was predictive enough. The closest model to our knowledge is Successor Representation \cite{Momennejad2017, Russek2017, Dayan1993}, that has been hypothesized to account for so-called model-free learning in cognitive systems \cite{Russek2017}. A major difference is that in Successor Representation, the value of a transition depends on the prediction error over the expected reward in the environment, whereas we assign value to a belief state via the probability of the associated sensory state in the absolute preference distribution. A limitation of our scheme, however, is that a target can only be encoded as a possibly ambiguous observation. Using reinforcement and a few interactions with the environment, the preference for a particular hidden state could be encoded in the look-ahead preference distribution in a hybrid scheme.

Conceptually, the FEPS framework fits into the field of NeuroAI \cite{Momennejad2023}, at the intersection of behavioral sciences, engineering and neurosciences. While Projective Simulation can be used to learn in an artificial environment, its vocation is to understand agency and the behavior of agents in the world. Furthermore, the FEPS attempts to give a biologically plausible account of learning and adaptive behavior, grounding internal computations in the active inference framework and the predictive processing paradigm. ECMs are potentially implementable on physical platforms and can be considered embodied structures underlying the memory of the agents. For example, a parallel can be drawn between the role the network of belief states plays for FEPS agents, and that of place cells and grid cells in the hippocampus \cite{Guntupalli2023, George2021, Whittington2022}. Both integrate stimuli to create a contextualized representation of an event in order to make predictions about future stimuli. For the FEPS to provide a modeling platform of interest for cognitive and behavioral sciences, the next challenge is to implement learning on real-world tasks and in a fully embodied way, including the calculation of coincidences between predictions and observations, and the update of the associations between states.

\section{Acknowledgments}{This research was funded in whole or in part by the Austrian Science Fund (FWF) [SFB BeyondC F7102, 10.55776/F71]. For open access purposes, the author has applied a CC BY public copyright license to any author accepted manuscript version arising from this submission. We gratefully acknowledge support from the European Union (ERC Advanced Grant, QuantAI, No. 101055129). The views and opinions expressed in this article are however those of the author(s) only and do not necessarily reflect those of the European Union or the European Research Council - neither the European Union nor the granting authority can be held responsible for them.}

\bibliographystyle{ieeetr}
\bibliography{FEPS_Biblio}

\appendix

\section{Expected free energy in the wandering phase} \label{appendix: EFE wandering IG}

During the wandering phase, the preference distribution is designed to encourage the agent to seek the relevant information to complete its model. The preferences are equal to the world model marginalized over actions:
\begin{equation}
 \text{pref}(S_{t+1},B_{t+1}|b_t) = \sum_{a} \pi(a|b_t)\, p(B_{t+1}|b_t, a)\, p(S_{t+1}|B_{t+1}).\end{equation}

Plugging this result in the expected free energy in Equation \eqref{eq: EFE}, one obtains:
\begin{align}
\mathcal{G}_{b_t}[a_t] = 
&  \mathbb{E}_ {b_{t+1},s_{t+1} \sim p(B_{t+1} , S_{t+1}| b_t, a_t)} [\log  p(b_{t+1} |b_t, a_t)
 - \log \text{pref}(s_{t+1},b_{t+1} | b_t,a_t)] \\
=  &   \sum_{b_{t+1},s_{t+1}} p(b_{t+1}|b_t, a_t)\, p(s_{t+1}|b_{t+1})\left[\log  p(b_{t+1} |b_t, a_t)
 - \log \sum_{a} \pi(a|b_t)\, p(b_{t+1}|b_t, a)\, p(s_{t+1}|b_{t+1})\right] \\
 = & \sum_{b_{t+1},s_{t+1}} p(b_{t+1}|b_t, a_t)\, p(s_{t+1}|b_{t+1}) \left[\log  p(b_{t+1} |b_t, a_t)
 - \log \sum_{a} \pi(a|b_t)\, p(b_{t+1}|b_t, a) - \log p(s_{t+1}|b_{t+1})\right] \\
 =& \sum_{b_{t+1}} p(b_{t+1}|b_t, a_t)  \left[- \sum_s p(s_{t+1}|b_{t+1})\log p(s_{t+1}|b_{t+1})\right] \\
 \notag & + \sum_{b_{t+1}} p(b_{t+1}|b_t, a_t) \left[\log  p(b_{t+1} |b_t, a_t)
 - \log \sum_{a} \pi(a|b_t)\, p(b_{t+1}|b_t, a)\right].
\end{align}

In the second line, we replace the preference distribution by its definition with respect to the world model, and the additivity rule of the logarithm is used to move to the third. In the fourth fourth line, we separated the expectation values over $B_{t+1}$ and $S_{t+1}$ by noticing that for two random variables $X,Y, \ \mathbb{E}_{x\sim p(X)}[f(Y=y)] = f(y)$ by the normalization constraint on probability distributions. As a result, terms are grouped by dependency, where in particular, the second term has become independent of sensory states. The first term is the conditional entropy of sensory states, conditioned on belief states. For some belief state $b_{t+1}$, and when the world model is clone-structured, the likelihood is a delta function: $p(s_{t+1}|b_{t+1}) = \delta_{s_{t+1}, s(b_{t+1})}$ where $s(b_{t+1})$ designates the observation $b_{t+1}$ is a clone of. As a result, when the sensory states match, the logarithm vanishes, and otherwise, it is multiplied by zero. Therefore, the entropy over observations cancels by design. The remaining term reduces to the Kullback-Leibler divergence between the transition function and the transition function marginalized over actions:
\begin{align}
\mathcal{G}_{b_t}[a_t] &= \mathbb{D}_{KL}\left[p(B_{t+1}|b_t, a_t) || \sum_{a} \pi(a|b_t)\, p(B_{t+1}|b_t, a) \right]\\
& = \mathbb{D}_{KL}\left[p(B_{t+1}|b_t, a_t) ||  p(B_{t+1}|b_t) \right] \\
&= \mathbf{IG}(B_{t+1}, A_t = a_t),
\end{align}
where $p(B_t|b_t)$ is the marginal of the transition function over actions, and in the last line, $\mathbf{IG}(X|Y=y)$ is the information gain about $X$ from knowing the value of $Y=y$.

\section{Limits of the EFE} \label{appendix: limit EFE wandering}

In this section, we derive, under certain assumptions that we justify with our numerical simulations, a limit for the expected free energy for environments with deterministic transitions, which include the timed response and navigation tasks. Formally, this means that the world model has been fully trained and the transition function yields perfect prediction accuracy over sensory states given the correct belief state is known. Equivalently, the variational free energy is minimized, and the posterior and prior distributions for the transition functions stay essentially identical after each update, i.e. $\exists \epsilon \in \mathbb{R}, \text{such that } \mathbb{D}_{\mathrm{KL}}[q_\phi(B_{t}|b_{t-1}, a_{t-1})|| p(B_{t}|b_{t-1}, a_{t-1})]< \epsilon$, where $\epsilon > 0$ can come arbitrarily close to 0. Then, considering the expression for the VFE in Eq. \eqref{eq: VFE}, we see that the first term can be neglected and the second term simplifies to $- \sum_{b_{t}} p(b_{t}|b_{t-1}, a_{t-1}) \log p(s_{t}^\mathrm{env}|b_{t}) $, where we used the fact that the Kullback-Leibler divergence cancels out if and only if the two distributions are equal, that is, $q_\phi(B_{t}|b_{t-1}, a_{t-1})= p(B_{t}|b_{t-1}, a_{t-1})$. In this limit, a good approximation of the VFE is: $\mathcal{F} = - \sum_{b_{t}} p(b_{t}|b_{t-1}, a_{t-1}) \log p(s_{t}^\mathrm{env}|b_{t}) = 0$ for any observation received from the environment. Furthermore, our assumption of a perfect world model implies that each belief state $b$ is associated to exactly one hidden state $e$ in the (deterministic) environment. That means that when the agent is in belief state $b$, the environment is in hidden state $e$. In these conditions, we say that $b$ \textit{represents} $e$. As a result, the transition function has the following form:
\begin{equation} \label{annex_eq: transition function}
p(b_{t}|b_{t-1},a_{t-1}) = \begin{cases} 
x_b  &\text{ if } s(b_{t}) = s_{t}^\mathrm{env} \text{ and } b_{t} \text{ represents a hidden state $e_{t}$ that emitted }s_{t}^\mathrm{env} \\
0 &\text{otherwise.}
\end{cases}
\end{equation}
such that $x_b \in [0,1]$ and the sum of probabilities for all belief states that represent the same hidden state $e_t$ sum up to 1: $\sum_b x_b = 1$. In principle, the values for the $x_b$ can be arbitrary, and depend on the individual models the agents collapse onto after training. We call \textit{children} of $b_t$ and $a_t$ the set of all belief states that can be reached with non-zero probability from $b_t$ with action $a_t$: $ch(b_t, a_t) = \{b_{t+1} | p(b_{t+1}|b_t,a_t) > 0 \}$. Similarly, we call $e(b_t,a_t)$ the hidden state in the environment that results from applying $a_t$ from $b_t$.  $e(b_t, a_t) = e_{t+1}$ designates the fact that starting from belief state $b_t$ and under action $a_t$, the environment will transition to $e_{t+1}$ in the next step.

Two asymptotic configurations stood out during the numerical simulations: (1) the agent adopts a distributed representation of the hidden state, and any belief state participating in it can be sampled with close to uniform probability (that we idealize to be exactly uniform in the derivation), or (2) reinforcements of individual edges in the models led to the adoption of a single belief state to fully represent the hidden state while ignoring the other clones associated with zero probability. A belief state represents at most one hidden state. In order to accommodate the first possibility, we define the set: \begin{equation}
\mathcal{D}_{b_t, e_{t+1}} = \{b_{t+1} | \exists a\in\mathcal{A}, \ p(b_{t+1}|b_t, a) > 0 \text{ and $b_{t+1}$ represents $e_{t+1}$ only}\}
\end{equation}
the set of all belief states that can be transitioned to and that uniquely represent the hidden state $e(b_t, a_t)$. Furthermore, we require that belief states $b_{t+1}, b'_{t+1} \in \mathcal{D}_{b_t, e_{t+1}}$ are degenerate, that is, $\forall a\in\mathcal{A}, \ p(b_{t+1}|b_t, a)=p(b_{t+1}'|b_t, a)$.  The size $|\mathcal{D}_{b_t, e}|$ of the set a belief state belongs to is called its \textit{degeneracy} and $\forall b \in \mathcal{D}_{b_t, e(b_t, a_t)}, \ p(b|b_t, a_t) = 1/|\mathcal{D}_{b_t, e(b_t, a_t)}|$. As a result, the world model becomes:
\begin{equation} \label{eq_annex: world model}
p(b_{t+1}, s_{t+1}|b_t, a_t) = \frac{\delta_{b_{t+1}\in ch(b_t, a_t)}}{|\mathcal{D}_{b_t, e(b_t, a_t)}|} \delta_{s_{t+1}, s(b_{t+1})},
\end{equation}
where as in the main text, $\delta_{b_{t+1}\in ch(b_t, a_t)}$ equals 1 if $b_{t+1}$ is a child of $b_t$ and $a_t$ and $0$ otherwise, and $s(b_{t+1})$ is the sensory state that $b_{t+1}$ is a clone of, in contrast to $s_{t+1}$, the value that the sensory state can take. Another important consequence is that if two actions $a, a'\in\mathcal{A}$ lead to the same hidden state in the environment, that is, $e_{t+1} = e(b_t, a) = e(b_t, a')$, then the children of $b_t$ and $a$, and $b_t$ and $a'$, respectively, are the same: $ch(b_t,a) = ch(b_t,a')$. Conversely, if $e(b_t, a) \neq e(b_t, a')$, then the two sets have no belief state in common: $ch(b_t,a) \cap  ch(b_t,a') = \emptyset$. Therefore, the following holds for any two actions $a, a'\in\mathcal{A}$:
\begin{equation} \label{eq_annex: equivalence D and children}
\exists b\in\mathcal{B} \text{ such that} \ b \in ch(b_t, a) \text{ and } b \in ch(b_t, a') \iff ch(b_t, a) = ch(b_t, a')
\end{equation}

Using Eq. \eqref{eq_annex: world model} for the world model, the new expression for the expected free energy (Eq.\eqref{eq: EFE} in the main text) is:
\begin{align}
\mathcal{G}_{b_t}[a_{t}] &= \sum_{b_{t+1}, s_{t+1}} \frac{\delta_{b_{t+1}\in ch(b_t, a_t)}}{|\mathcal{D}_{b_t, e(b_t, a_t)}|} \delta_{s_{t+1}, s(b_{t+1})} \log\left(\frac{\delta_{s_{t+1}, s(b_{t+1})}  / |\mathcal{D}_{b_t, e(b_t, a_t)}|}{\text{pref}(b_{t+1},s_{t+1}|b_t)}\right)\\
&= \sum_{b_{t+1}\in ch(b_t, a_t), s_{t+1}} \frac{1}{|\mathcal{D}_{b_t, e(b_t, a_t)}|} \delta_{s_{t+1}, s(b_{t+1})} \log\left(\frac{1  / |\mathcal{D}_{b_t, e(b_t, a_t)}|}{\text{pref}(b_{t+1},s_{t+1}|b_t)}\right).
\end{align}

\textbf{Wandering phase:}
During the wandering phase, the preferences of the agents are set as the marginal of their world model over actions:
\begin{align}
\text{pref}(b_{t+1},s_{t+1}|b_t) 
&= \sum_{a} p(s_{t+1}|b_{t+1})\, p(b_{t+1}|b_t,a) \,\pi(a|b_t) \\
&= \sum_{a} \delta_{s_{t+1}, s(b_{t+1})} \frac{\delta_{b_{t+1} \in \text{ch}(b_t,a)}}{|\mathcal{D}_{b_t, e(b_t, a)}|}\pi(a|b_t)
\end{align}
and the EFE is determined by the final policy:
\begin{equation}
\mathcal{G}_{b_t}[a_{t}] = \sum_{b_{t+1} \in\text{ch}(b_t,a_{t})} \frac{1}{|\mathcal{D}_{b_t, e(b_t, a)}|}\log \left( \sum_{a | e(b_t,a) = e(b_t, a_t)} \frac{1  / |\mathcal{D}_{b_t, e(b_t, a_t)}|}{\pi(a|b_t) \cdot1  / |\mathcal{D}_{b_t, e(b_t, a)}|}\right),
\end{equation}
where by \eqref{eq_annex: equivalence D and children}, the sum over actions in the logarithm is restricted to those whose children include $b_{t+1}$, or equivalently, represent the same hidden state and belong to the same set of degenerate states. Therefore, the degeneracies inside the logarithm cancel each other out, and each term in the sum is the same for each degenerate belief state:
\begin{equation} \label{eq_annex: EFE_wandering}
\mathcal{G}_{b_t}[a_{t}] = -\log \left( \sum_{a | e(b_t,a) = e(b_t, a_t)} \pi(a|b_t)\right).
\end{equation}

In order to find the final value the EFE converges to, we must find the fixed point for the policy. This means that the policy $\pi^{(n+1)}$ after the $n$-th update using the expected free energy is equal to the original policy at step $n$, $\pi^{(n)} = \pi$:
\begin{align}
\pi(a_{t}|b_t) &= \pi^{(n)}(a_{t}|b_t) \\
&= \pi^{(n+1)}(a_{t}|b_t)\\
&= \text{softmax}(\zeta \mathcal{G}_{b_t}[a_{t}]) \\
& = \frac{\exp \left(-\zeta \log \left( \sum_{a | e(b_t,a) = e(b_t, a_t)} \pi(a|b_t)\right)\right)}
{\sum_{a'} \exp \left(-\zeta\log \left( \sum_{a'' | e(b_t,a'') = e(b_t, a')} \pi(a''|b_t)\right)\right)} \\
&= \frac{\left( \sum_{a | e(b_t,a) = e(b_t, a_t)} \pi(a|b_t)\right)^{-\zeta}}
{\sum_{a'} \left( \sum_{a'' | e(b_t,a'') = e(b_t, a')} \pi(a''|b_t)\right)^{-\zeta}} \label{eq_annex: EFE_wander_1}
\end{align}
where we have used $\exp(y \log x) = (\exp(\log x))^y = x^y$ to get the last line and $b'_{t+1}$ in the denominator is a belief state that can be reached from $b_t$ with action $a'\in\mathcal{A}$. 

To proceed further, we introduce the following notations: \begin{itemize}
\item $\mathcal{E}_{b_t} := \{e_{t+1}^k \ |\ \exists a \in\mathcal{A} \text{ such that } e(b_t, a) = e_{t+1}^k\}$, the set of hidden states that can be reached from the belief state $b_t$;
\item $\mathcal{A}_{b_t, k} := \{a \ | e(b_t, a) = e_{t+1}^k,\ a\in \mathcal{A}\} $, the set of actions that induce a transition to the $k$-th hidden state $e_{t+1}^k$ in $\mathcal{E}_{b_t}$. $|\mathcal{A}_{b_t, k}|$ is the number of elements in $\mathcal{A}_{b_t, k}$. It has at least one element;
\item $\forall e_{t+1}^k \in \mathcal{E}_{b_t}, \forall a\in \mathcal{A}_{b_t, k},\ \pi_{b_t,k} = \pi(a|b_t)$ denotes the probability of taking any action $a$ that will result in the $k$-th hidden state $e_{t+1}^k$ in $\mathcal{E}_{b_t}$. All actions that result in the same hidden state $e_{t+1}^k$ have the same probability in the policy, as can be seen in Eq. \eqref{eq_annex: EFE_wander_1}.
\end{itemize}

We rewrite Eq. \eqref{eq_annex: EFE_wander_1} with this notations, assuming $a_t\in \mathcal{A}_{b_t, k}$:
\begin{align}
\pi(a_{t}|b_t) 
&= \pi_{b_t, k} \\
&= \frac{\left( |\mathcal{A}_{b_t, k}| \pi_{b_t, k}\right)^{-\zeta}}
{\sum_{l} |\mathcal{A}_{b_t, l}| \,\left( |\mathcal{A}_{b_t, l}| \pi_{b_t, l}\right)^{-\zeta}}  \\
&= \frac{|\mathcal{A}_{b_t, k}|^{-\zeta/(1+\zeta)}}
{\left(\sum_{l} |\mathcal{A}_{b_t, l}|\,\left( |\mathcal{A}_{b_t, l}| \pi_{b_t, l}\right)^{-\zeta}\right)^{1/(1+\zeta)}}  \\
&= N \times |\mathcal{A}_{b_t, k}|^{-\zeta/(1+\zeta)}
\end{align}
where from the second to third line, we used that for $x\neq 0,$ $x=yx^\alpha \iff x^{1-\alpha} = y$ , and $N$ is the normalization factor in the second to last line. $N$ does not depend on the action considered and it normalizes the policy to 1:
\begin{equation}
\sum_k |\mathcal{A}_{b_t, k}|\, \pi_{b_t, k} = N \times \sum_k |\mathcal{A}_{b_t, k}|^{1/(1+\zeta)} = 1.
\end{equation}
Therefore, we find $N = 1/\sum_k |\mathcal{A}_{b_t, k}|^{1/(1+\zeta)}$. At the end of a successful training in a deterministic environment,  the policy converges to:
\begin{equation}
\pi_{b_t,k} = \frac{|\mathcal{A}_{b_t, k}|^{-\zeta/(1+\zeta)}}{\sum_l |\mathcal{A}_{b_t, l}|^{1/(1+\zeta)}}.
\end{equation}
Finally, during the wandering phase, once the world model perfectly represents the environment, the expected free energy in Eq. \eqref{eq_annex: EFE_wandering} related to taking an action $a_t\in\mathcal{A}_{b_t,k}$ converges to:
\begin{align}
\mathcal{G}_{b_t}[a_t] 
&= - \log \left( \sum_{a' \in \mathcal{A}_{b_t,k}} \frac{|\mathcal{A}_{b_t, k}|^{-\zeta/(1+\zeta)}}{\sum_l |\mathcal{A}_{b_t, l}|^{1/(1+\zeta)}}\right) \\
&= - \log \left( \frac{|\mathcal{A}_{b_t, k}|^{1/(1+\zeta)}}{\sum_l |\mathcal{A}_{b_t, l}|^{1/(1+\zeta)}}\right).
\end{align}

In figures \ref{fig: DR} and \ref{fig: GW}, we average this value over actions and previous belief states, since it would otherwise depend on individual deliberations of the agents.

\section{Simulations parameters}

\begin{table} [h!]
    \centering
    \begin{tabular}{|c|c|c|} \hline 
         Parameters&  Skinner box& Navigation\\ \hline 
         $N_\text{agents}$&  100&30\\ \hline 
         $N_\text{clones}$& 2&3\\ \hline 
         $N_\text{episodes}$&  4 000& 40 000\\ \hline 
         Length episodes&  \multicolumn{2}{|c|}{80}\\ \hline 
         Forgetting rate $\gamma$&  \multicolumn{2}{|c|}{0.001}\\ \hline
         Reward scale R&  \multicolumn{2}{|c|}{3}\\ \hline 
         scaling parameter $\zeta_\text{wandering}$&  \{0\} & \{-3, -1, 0, 1, 3\}\\ \hline 
         scaling parameter $\zeta_\text{task}$&   -1& -3\\ \hline 
         p*&  \multicolumn{2}{|c|}{0.99}\\ \hline
 $N_\text{pref}$& \multicolumn{2}{|c|}{1}\\\hline
 Prediction horizon $T_h$& 2 & 3\\\hline
    \end{tabular}
    \caption{Parameters used during the training and the testing of the agents}
    \label{tab: simulation parameters}
\end{table}

\end{document}